\documentclass[10pt,twocolumn,letterpaper]{article}

\usepackage{iccv}
\usepackage{times}
\usepackage{epsfig}
\usepackage{graphicx}
\usepackage{amsmath}
\usepackage{amssymb}

\usepackage{tikz}
\usepackage{color,soul}

\usepackage{bbding}
\usetikzlibrary{calc}
\usepackage[export]{adjustbox}
\usepackage{gensymb}
\usepackage{pgfplots}

\usepackage[pagebackref=true,breaklinks=true,letterpaper=true,colorlinks,bookmarks=false]{hyperref}

\usepackage[capitalize]{cleveref}
\crefname{section}{Sec.}{Secs.}
\Crefname{section}{Section}{Sections}
\Crefname{table}{Table}{Tables}
\crefname{table}{Tab.}{Tabs.}

\iccvfinalcopy 

\def\iccvPaperID{8606} 

\ificcvfinal\fi

\begin{document}

\title{ORTexME: Occlusion-Robust Human Shape and Pose via \\Temporal Average Texture and Mesh Encoding}
\author{Yu Cheng$^1$, Bo Wang$^2$, Robby T. Tan$^{1}$\\
$^1$ National University of Singapore \hspace{3mm} $^2$CtrsVision \hspace{1mm}\\
\tt\small  e0321276@u.nus.edu, hawk.rsrch@gmail.com,
\tt\small  robby.tan@nus.edu.sg
}

\maketitle
\ificcvfinal\thispagestyle{empty}\fi

\begin{abstract}
    In 3D human shape and pose estimation from a monocular video, models trained with limited labeled data cannot generalize well to videos with occlusion, which is common in the wild videos. 
    The recent human neural rendering approaches focusing on novel view synthesis initialized by the off-the-shelf human shape and pose methods have the potential to correct the initial human shape. 
    However, the existing methods have some drawbacks such as, erroneous in handling occlusion, sensitive to inaccurate human segmentation, and ineffective loss computation due to the non-regularized opacity field. 
    To address these problems, we introduce ORTexME, an occlusion-robust temporal method that utilizes temporal information from the input video to better regularize the occluded body parts.  
    While our ORTexME is based on NeRF, to determine the reliable regions for the NeRF ray sampling, we utilize our novel average texture learning approach to learn the average appearance of a person, and to infer a mask based on the average texture.
    In addition, to guide the opacity-field updates in NeRF to suppress blur and noise, we propose the use of human body mesh. 
    The quantitative evaluation demonstrates that our method achieves significant improvement on the challenging multi-person 3DPW dataset, where our method achieves 1.8 P-MPJPE error reduction. The SOTA rendering-based methods fail and  enlarge the error up to 5.6 on the same dataset.
    %
    %

\end{abstract}

\section{Introduction}

\begin{figure}[t]
   \centering
   \includegraphics[width=1.0\linewidth]{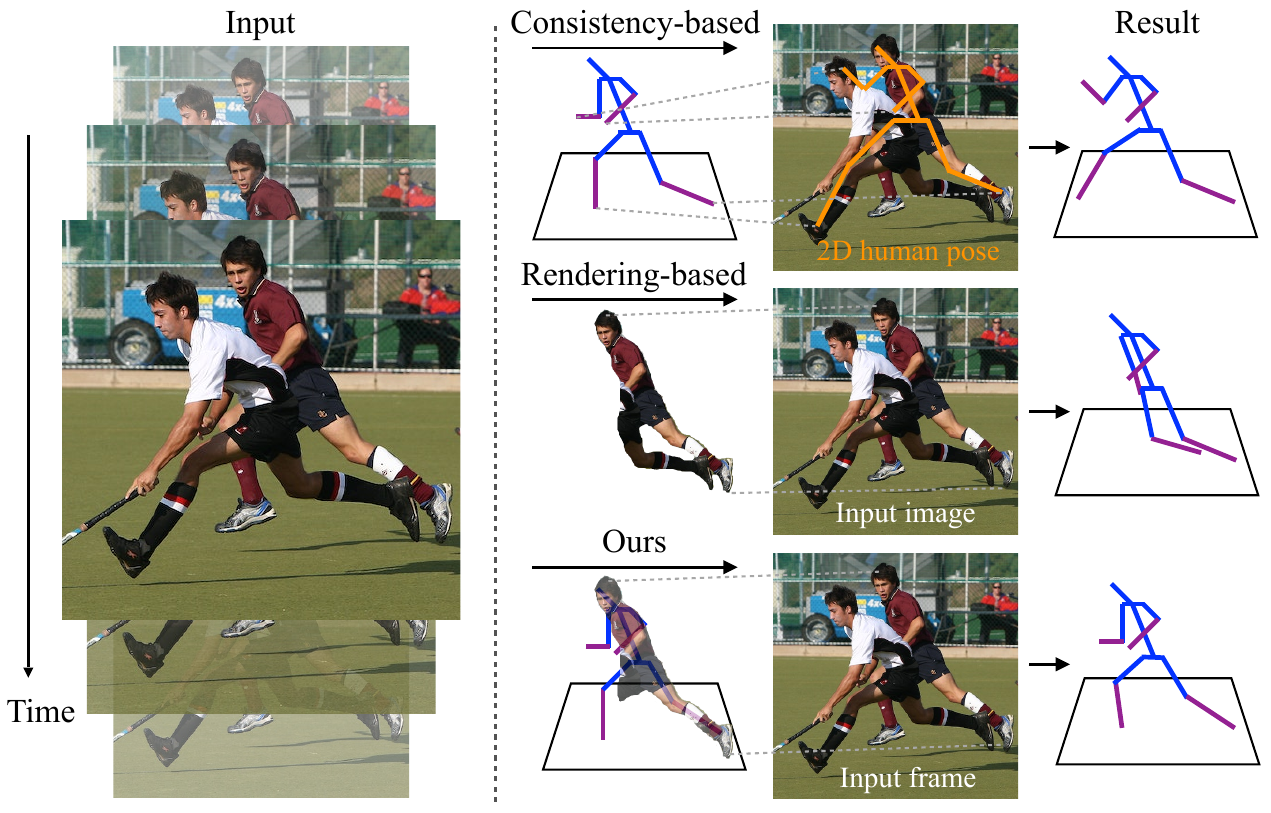}
   \caption{Conceptual illustration of the differences between the consistency-based, rendering-based, and our proposed methods. First row: the consistency-based correction approaches try to match to the 2D human pose estimation, which is affected by the 2D pose errors. Second row: the rendering-based approaches render a target person based on the initial human shape and pose, which is sensitive to occlusion and wrong segmentation. Last row: our approach is robust to occlusion by taking the segmentation error into consideration and leveraging the temporal information.}
   \label{fig:concept}
\end{figure}

3D human shape and pose estimation is a fundamental computer vision problem and serves as a building block for many downstream tasks (e.g,~\cite{yan2018spatial,shi2019two,liu2020disentangling,qian2018pose,li2019cross,miao2019pose,pereira2019fast,li2020deformation}). 
%
Many supervised methods have been proposed to estimate 3D human shape and pose from monocular image or video (e.g., \cite{bogo2016keep,kanazawa2018end,kolotouros2019spin,humanMotionKanazawa19,kocabas2020vibe,choi2021beyond,kocabas2021pare,rempe2021humor,khirodkar2022occluded}). 
%
However, due to the costly ground truths, only few video datasets with labels are available for training, limiting the generalization ability of the supervised methods on the wild videos.

To reduce the dependency on the training data, self-supervised losses for 3D human shape and pose estimation have been developed, such as enforcing the temporal smoothness~\cite{zeng2022smoothnet,cheng2022dual}, applying the consistency between different rotations with virtual camera \cite{chen2019unsupervised,kocabas2019self,zhang2020inference}, or measuring the consistency between 2D and 3D human poses \cite{tung2017self,wandt2019repnet,cheng2022dual}. However, one drawback of consistency-based approaches is that the 2D human poses used can be incorrect, particularly in occlusion scenarios, as illustrated in the first row of \cref{fig:concept}. 

Recent NeRF-based novel view synthesis~\cite{kwon2021neural,peng2021neural,weng2022humannerf} has been shown to benefit the 3D human shape and pose model~\cite{su2021nerf,weng2022humannerf,te2022neural}. 
Using view synthesis, we can compare of the rendered images directly with the input images, instead of certain pseudo labels such as 2D human poses. 
Unfortunately, the existing rendering-based methods \cite{su2021nerf,weng2022humannerf,te2022neural} cannot handle occlusions (i.e., self-occlusions, object occlusions, and inter-person occlusions). 
Occlusions cause the pixel-wise RGB loss in the occluded body parts to be incorrect. 
The existing methods also rely on an accurate human segmentation mask for NeRF ray sampling. Thus, they are sensitive to the segmentation inaccuracy, which is unavoidable for videos in the wild, particularly in multi-person scenarios, as illustrated in the second row of \cref{fig:concept}. 
Moreover, the existing methods do not have effective regularization for NeRF's opacity field, leading to a blurry opacity field and an ineffective image-level loss. 

To address the problems, we introduce \emph{ORTexME}, an occlusion-robust temporal method  to correct the errors in 3D human pose and shape estimation. 
First, our ORTexME exploits temporal information from the input video frames by ensuring the predicted human poses follow the human dynamics of the visible human body parts. 
In particular, we incorporate a number of frames' feature encodings as input, which contains the dynamics of the human pose sequence.
Second, the existing rendering-based methods~\cite{su2021nerf,weng2022humannerf,te2022neural} rely on an accurate human segmentation mask, and assume its availability. 
However, segmentation masks can fail to include some parts of a target person, or can wrongly include another person who is occluding the target person in multi-person scenarios. 
Hence, we propose an average texture learned from the video sequence to infer a mask, and thus avoid severe segmentation errors.

Third, compared with A-NeRF~\cite{su2021nerf}, where only skeletal information is used, our human body mesh-based approach also considers the shape coefficients of the target person.
The introduction of the human body shape helps to regularize the opacity field to be sharper and less noisy, enabling the rendered persons to be clearer and the regressed 3D human poses to be more accurate.

In summary, our major contributions are as follows:
\begin{itemize}
    \vspace{-0.2cm}
    \item We propose a novel method that utilizes temporal information to handle occlusions by ensuring the proper regularization of the invisible human body parts of a 3D human pose. We embed this temporal information into a NeRF-based network.
    \vspace{-0.2cm}
    \item We introduce a novel average texture learned in a canonical space to have occlusion-robust masks. This is particularly important in multi-person settings. 
    \vspace{-0.2cm}
    \item We present a human body mesh-based encoding to generate a clean opacity field for an improved rendered person, resulting in effective loss computation for 3D human pose refinement. 
    \vspace{-0.2cm}
\end{itemize}
To the best of our knowledge, we are the first to achieve positive error reduction among the rendering-based methods on the multi-person 3DPW dataset, where we reduce \textbf{1.8} P-MPJPE error compared with the SOTA methods \cite{su2021nerf,weng2022humannerf} where they instead enlarge the error up to 5.6. 
%


\begin{figure*}[h!]
	\centering
	\includegraphics[width=\textwidth]{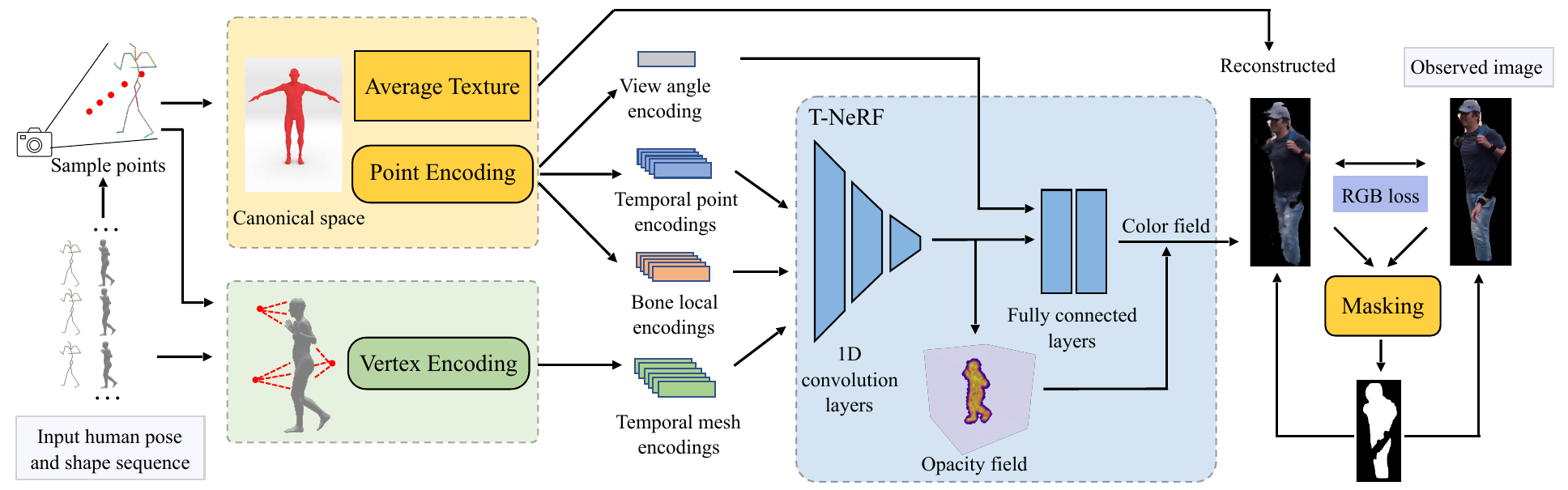}
	\caption{An overview of our framework. Each of the 3D human shape and pose sequences from an off-the-shelf method is used to sample points, followed by point and vertex encodings. The average texture of the current person is learned in the canonical space for predicting a mask of the reliable region to avoid wrong human segmentation. The vertex encoding helps to regularize NeRF's opacity field to be sharp and clean. The encodings of the input sequence are fed into a temporal NeRF to make use of temporal information to handle occlusions when no visible information is available for the occluded body parts.}
	\label{fig:overview}
	\vspace{-0.8em}
\end{figure*}

\section{Related Work}

\noindent\textbf{3D Human Shape Estimation from Videos} 
Supervised methods for 3D human pose or shape estimation from monocular videos have been developed to make use of temporal cues \cite{zhou2016sparseness,mehta2017vnect,pavllo20193d,humanMotionKanazawa19,cheng2019occlusion,kocabas2020vibe,choi2021beyond}, which demonstrate improved performance compared with their counterparts that take single images as input \cite{martinez2017simple,chen20173d,li2019generating,kolotouros2019learning,moon2020i2l}. 
However, the availability of public video datasets is limited due to the high cost of obtaining 3D human pose ground truth. Consequently, models trained on these datasets may have limited generalization ability on wild videos and exhibit low cross-dataset performance.

\noindent\textbf{Consistency-based 3D Human Modeling} 
In the multi-view setting, self-supervised approaches using multi-view geometry \cite{kocabas2019self} or learnable triangulation \cite{iskakov2019learnable} can be used, but multi-view input is required during training or testing \cite{pavlakos2017harvesting}. 
In the monocular scenario, 
consistency losses are proposed to check if the 2D projection of random rotation or transformation of the 3D pose is consistent with the original 2D pose \cite{chen2019unsupervised,zhang2020inference}. 
Checking between the 3D and 2D poses are developed to use a reprojection layer to project 3D pose back to 2D with explicit camera parameter estimation \cite{wandt2019repnet}, check the consistency between reprojected 3D and 2D human poses \cite{cheng2022dual}, or combine body joints, segmentation, and mesh vertex motion together \cite{tung2017self}. However, these approaches rely on pseudo labels like 2D poses, which cannot directly refine the 3D poses from the observation.

\vspace{0.1cm}
 \noindent\textbf{Rendering-based 3D Human Modeling} 
NeRF is proposed for continuous neural scene representation from multi-view images, which demonstrates photorealistic results of novel-view synthesis but assumes a scene is static without objects moving \cite{mildenhall2020nerf}. To extend NeRF for dynamic scenes or moving/deforming objects, dynamic NeRFs \cite{pumarola2021d,park2021nerfies,attal2021torf} are proposed, where the image from each view is mapped to a common canonical space. However, the dynamic NeRF methods are general but not specifically designed for 3D human modeling. Recent works explore extending NeRF to 3D human modeling \cite{su2021nerf,weng2022humannerf,te2022neural}, but these methods cannot handle occlusions because no effective loss can be computed in the invisible region where human body is occluded. Moreover, these methods are sensitive to inaccurate human segmentation because non-target person texture would be used in the loss computation for the target person, leading to the NeRF-based human rendering being erroneous. 
Last, the existing methods do not provide appropriate constraints in NeRF's opacity field,
resulting in blurring rendering quality, and affecting the image-level loss computation.

\section{Proposed Method}

The overview of our framework is shown in \cref{fig:overview}. 
Given an initial 3D human pose and shape sequence, each human pose is used to sample points in the observation space, where the points are encoded in two ways. 
The first encoding is point encoding in the canonical space, in which the average texture of the current person is learned. 
The second encoding is vertex encoding, which helps to regularize NeRF's opacity field to be less blurry.
Both the point and vertex encodings of the input pose sequence are the input of our temporal NeRF network.

Compared with the existing frame-by-frame approach~\cite{su2021nerf}, our temporal NeRF is capable of handling occlusions (i.e., invisible human body parts) by exploiting the temporal information. 
Finally, benefiting from our average texture learned in the canonical space, we can dynamically obtain a mask at each frame (excluding the unreliable regions caused by incorrect human segmentation) by comparing the average texture-based rendering of a target person to the observation from the input image. 
This strategy is effective in handling wrong segmentation, which happens frequently in multi-person settings. 

\subsection{Point Encoding}
\label{sec:point_encoding}

\begin{figure}
    \centering
    \includegraphics[width=0.88\linewidth]{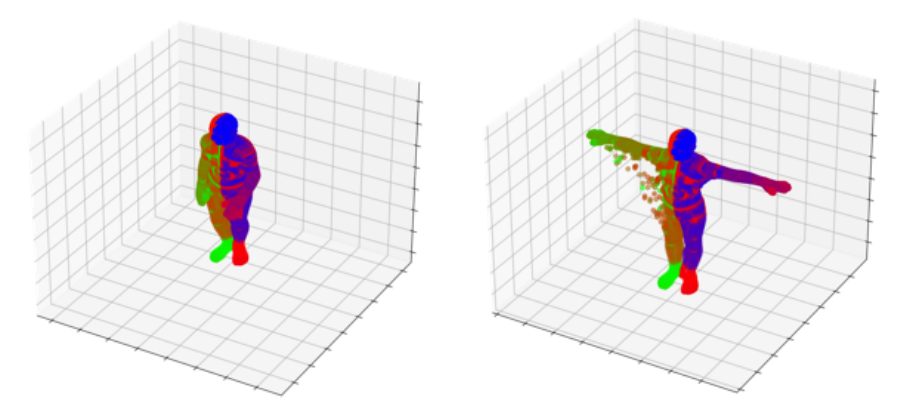}
    \caption{$R^3 \xrightarrow[]{} R^3$ mapping introduces ambiguity for overlapping mesh points. Left: Limbs of a target person are overlapped in the observation space. Right: the sample points that overlap become distorted in the canonical space. 
    }
    \label{fig:can_space}
\end{figure}

Mapping a person from the observation space to the canonical space has been explored recently \cite{weng2022humannerf,te2022neural,li2022tava}, where mapping functions in both directions are proposed. 
Specifically, they utilize a $R^{3} \rightarrow R^{3}$ mapping, which can warp the body parts correctly when they are separated. 
However, as shown in \cref{fig:can_space}, the mapping fails to transform the body parts to the canonical space when the body surfaces are attached to each other (e.g., arm to torso), and may get worse when the 3D human pose is inaccurate.

Instead of using the $R^{3} \xrightarrow{} R^{3}$ mapping, we propose to use a less restricted canonical-space mapping: $R^{3} \xrightarrow{} R^{3B}$, where $B$ is the number of bones in the skeleton. 
Combined with the temporal information (detailed in \cref{sec:temporal}), our network learns the motion of each bone separately, instead of treating the body as a whole, 
and thus reduces the ambiguity caused by overlapping meshes. Specifically, our encoding is expressed as:
\begin{equation}
    p(x) = [(R_{1}^{-1}x-T_{1}), (R_{2}^{-1}x-T_{2}), ..., (R_{B}^{-1}x-T_{B})], 
\end{equation}

\noindent where $x$ is the coordinate of a sample point in the observation space. $R_{b}$ and $T_{b}$ are the rotation and translation matrices of each bone from the canonical space to the observation space, where the transformation is commonly used in the literature {\cite{weng2022humannerf, su2021nerf, te2022neural}}.
The final encoding $p$ is the concatenation of points in the canonical space with respect to each bone.

Similarly, for viewing angles, we also encode the viewing angle vector with $R$ and $T$:
\begin{equation}
    w(v) = [(R_{1}^{-1}a-T_{1}), (R_{2}^{-1}a-T_{2}), ..., (R_{B}^{-1}a-T_{B})],
\end{equation}
\noindent where $a$ is the viewing angle vector from the camera center to a particular pixel in the image plane.

\subsection{Vertex Encoding}
\label{sec:vertex_encoding}

\begin{figure}[tb]
    \centering
    \includegraphics[width=\linewidth]{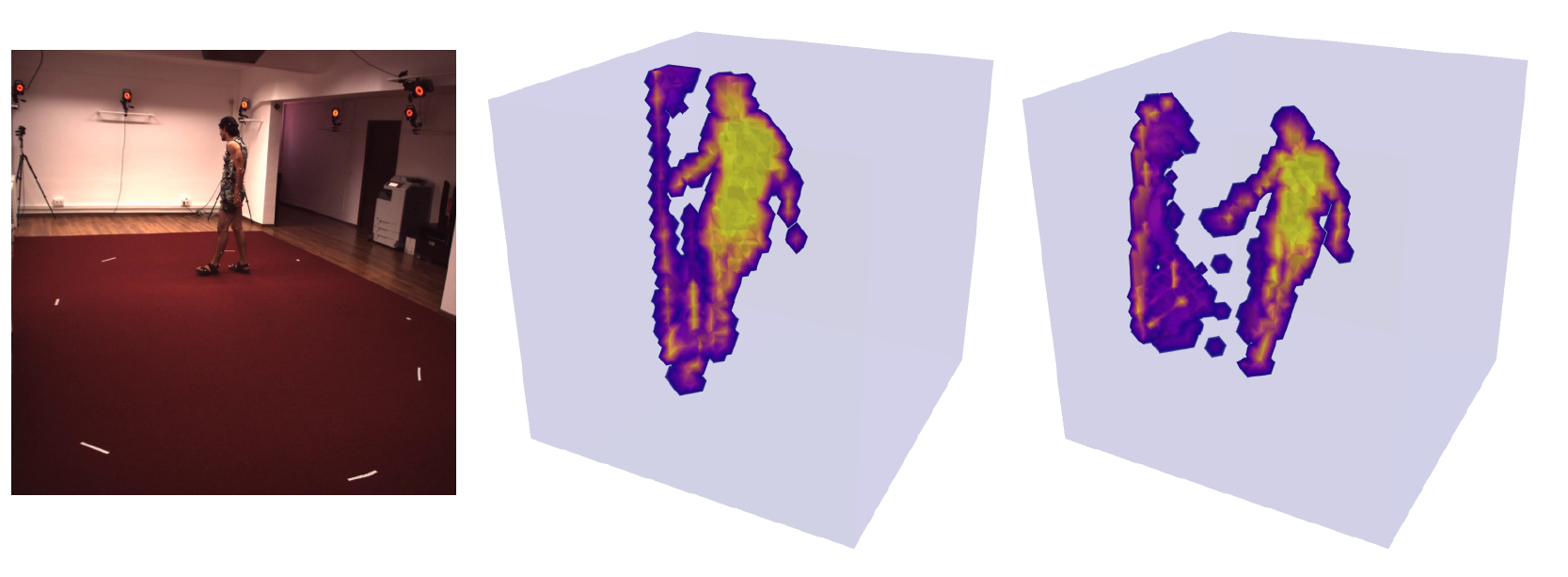}
    \caption{Effectiveness of the proposed vertex encoding with the obtained opacity field. Left: original image; Middle: Opacity field without vertex encoding, where the foreground and background are fused; Right: Opacity field with vertex encoding, where opacity field is sharper and better separated from the background.}
    \label{fig:opacity}
\end{figure}

Besides the point and viewing angle encoding, the human mesh can help the NeRF human reconstruction.
The human body surface can be used as a strong constraint for the human opacity, where the opacity should be high around the human body mesh and low for locations far from the mesh. 
Our vertex encodings can effectively constrain the opacity field to be less blurry, as shown in \cref{fig:opacity}. 
Unlike the latent encoding that only includes a limited distance around the human mesh \cite{te2022neural,peng2021neural},
we dynamically merge the latent vertex vectors in the whole observation space from the original image. 
Specifically, we use a Gaussian distance as the weight to compute the weighted sum of the latent vectors of a given sample point's location $x$. 
\begin{equation}
    v(x) = \sum_{k=1}^{K} \exp(- \frac{(x-u_{k})^2}{2s^2}) l_{k},
\end{equation}
\noindent where $K$ is the number of the K-nearest neighbors and $l_{k}$ is the latent vector of vertex $u_{k}$. $s$ is a pre-defined standard deviation,
%
which is the parameter to control the coefficient of the Gaussian distance.

\subsection{Temporal NeRF}
\label{sec:temporal}

With the two encodings from \cref{sec:point_encoding} and \cref{sec:vertex_encoding}, 
we obtain the representation of the sample points in the observation space according to the input human pose sequence. 
Instead of feeding a single frame's encodings into NeRF \cite{su2021nerf,weng2022humannerf,te2022neural}, we incorporate multiple frame's information as the input for our temporal NeRF, which is illustrated in the right part of \cref{fig:overview} (blue block). 

Our temporal NeRF contains two parts: temporal opacity and color field networks. 
The temporal opacity network consists of a number of TCN (temporal convolutional network) layers, which takes multiple frames' point and vertex encodings as input. 
These encodings provide the information of the pose sequence, which intrinsically contains the dynamics of human poses.
Thus, they help to deal with occlusion, where the invisible body parts can be inferred from the temporal information of the neighboring frames. 
Similar to the existing NeRF structure, our temporal network receives only the point and vertex encodings to generate the opacity field $\sigma$ due to the view-independency. 
Mathematically, the generation process is expressed as:
\begin{equation}
    \sigma_{i} = f_{\rm temp} (p_{i,1...T}, v_{i,1...T}),
\end{equation}
\noindent where $p_i$ and $v_i$ are the point and vertex encodings of the $i^{th}$ sample point in the observation space.

Different from the view-independency of the opacity field, the rendered colors are related to the viewing angles.
Thus, our color field network receives the intermediate features of our temporal opacity network, and views the encodings as the input to generate the color field $c$, which is defined as:
\begin{equation}
    c_{i} = f_{\rm color} (f_{\rm temp}(p_{i,1...T}, v_{i,1...T}), w_{i,1...T}).
\end{equation}
Finally, the RGB value $C$ of a given pixel via the volumetric rendering is achieved by:
\begin{eqnarray}
    C &=& \sum_{i} {\rm Tr}_i(1-\exp(-\sigma_{i}\delta_{i})) c_{i},\\
    {\rm Tr}_i &=& \exp(-\sum_{j=1}^{i-1}\sigma_{j}\delta_{j}),
\end{eqnarray}
\noindent where $\delta_{i}$ is the depth difference between two sample points. 

\subsection{Average Texture-based Masking}
\label{sec:masking}

\begin{figure}
    \centering
    \includegraphics[width=0.96\linewidth]{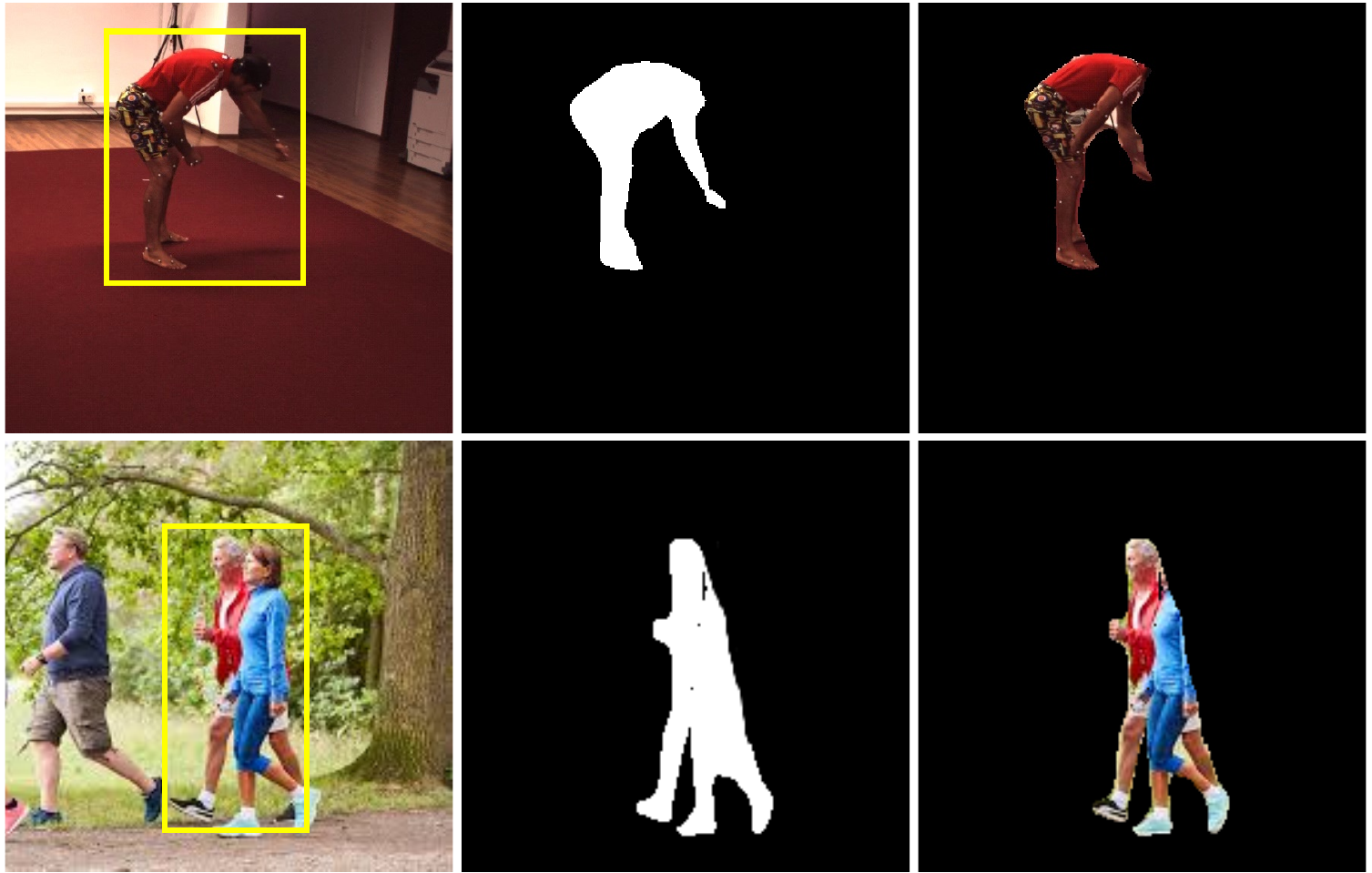}
    \caption{Examples of incorrect human segmentation where the target person is highlighted in the yellow rectangle. First row: the segmentation is incomplete due to the challenging foreground and background. Second row: the segmentation erroneously includes the person with the blue shirt in front.}
    \label{fig:segerr}
\end{figure}

Human segmentation can be inaccurate due to challenging image appearance or inter-person occlusions. 
As shown in \cref{fig:segerr}, if the segmentation of a person is wrong (middle column), the image reconstruction of the person using NeRF becomes erroneous (right column).
The existing NeRF-based 3D human pose methods \cite{su2021nerf, te2022neural} assume that an accurate segmentation mask is available, which is used to sample the pixels for RGB loss. When the inaccurate segmentation mask is provided, the background or non-target object is included as noise for the loss computation. Thus, these methods are sensitive to the segmentation inaccuracy. 

\begin{figure}
    \centering
    \includegraphics[width=0.93\linewidth]{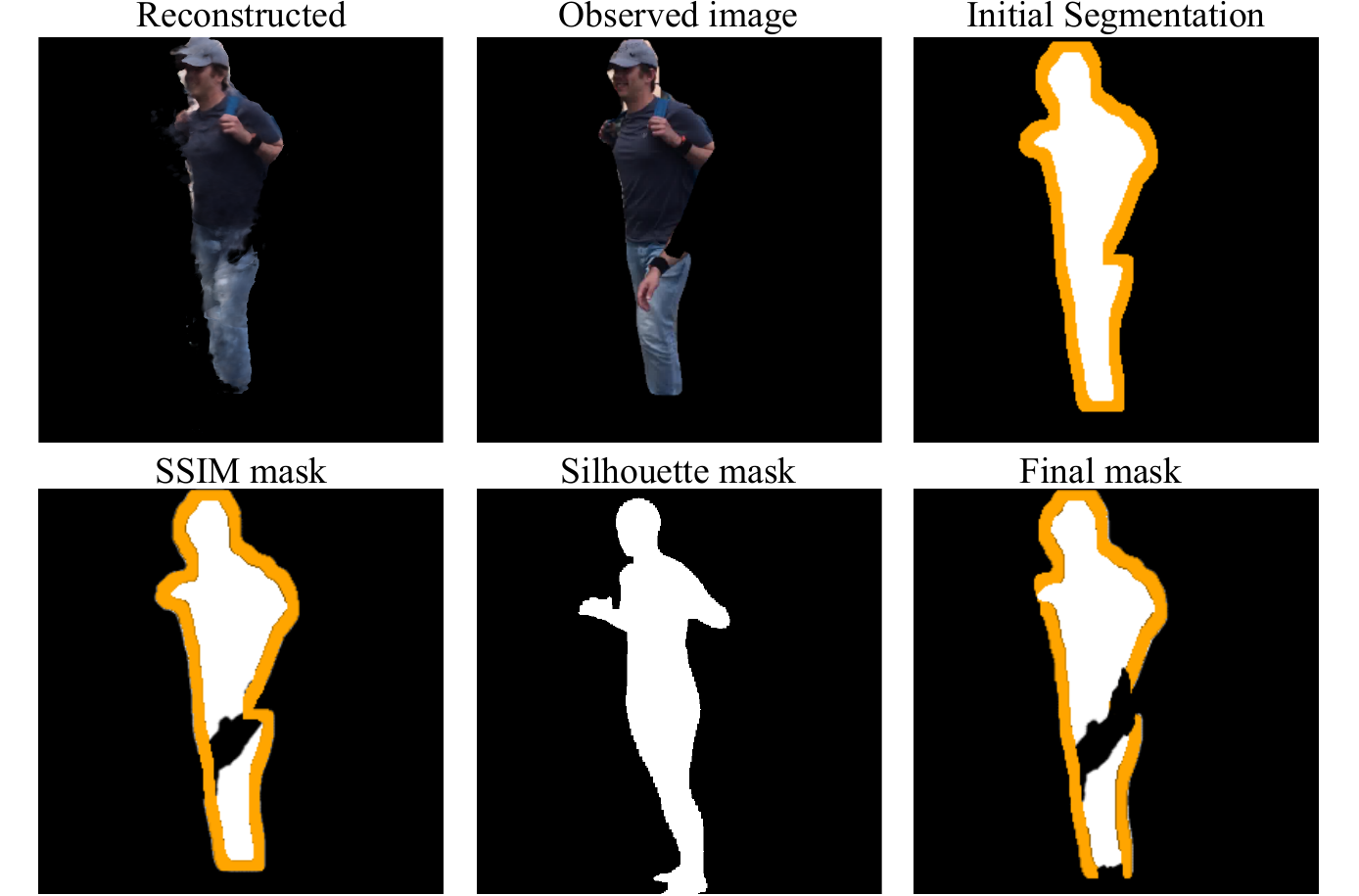}
    \caption{The procedure of the average texture-based masking where segmentation and zero-padding are colored as white and orange, respectively. First, we compute the SSIM between the reconstructed image and the observed image, and the less similar areas are excluded. Then, the silhouette mask derived from pose and shape via SMPL model \cite{loper2015smpl} is utilized to refine the SSIM mask to obtain the final mask.}
    \label{fig:probmask}
\end{figure}

To mitigate the sensitivity issue regarding segmentation errors, we estimate masks for each frame from reliable areas.
The procedure is illustrated in \cref{fig:probmask}. 
First, based on the current frame's pose $\theta$ and shape $\beta$, we obtain the reconstructed image $I_r$ via NeRF and compute the structural similarity index measure (SSIM) between the reconstructed image $I_r$ and observed image $I_o$ to obtain the SSIM mask $M_{\rm SSIM}$. The mask is computed using the Otsu method~\cite{otsu1979threshold} formulated as:
\begin{equation}
    M_{\rm SSIM} = {\rm Otsu}({\rm SSIM}(I_o, I_r(\theta, \beta))),
\end{equation}

The SSIM mask measures the difference between the reconstructed image and the observed image, where similar areas are regarded as reliable regions to train our network. 
The inconsistent regions of the initial segmentation mask are typically non-target objects such as occluders, as illustrated in \cref{fig:probmask}. 
In this example, the hand of a nearby person occludes the target person in the observed image, and the hand is included as part of the initial segmentation mask. 
In contrast, our predicted SSIM mask is able to exclude the inconsistent regions caused by occlusion.


Second, we compute the human body silhouette $S_h$ as an additional mask to determine the non-zero areas (orange color in the initial segmentation of \cref{fig:probmask}). 
Assuming the border of segmentation is padded with zero values $P_0$, the silhouette mask can filter the invalid zero-padding with the pose and shape information as shown in \cref{fig:probmask}, where the invalid padding within the orange area is removed according to the silhouette mask. The final mask $M$ is formulated as:
\begin{equation}
    M = M_{\rm SSIM} (1 - S_h P_0).
\end{equation}

Lastly, to ensure each frame contains enough information to refine the human shape and pose, certain frames may be skipped from the RGB loss computation as in \cref{eq:RGB_loss}. In case the obtained final mask of a frame is too deviated from the initial segmentation mask, the current frame is excluded in the downstream processing, because the non-occluded human body can be too small or the segmentation error is too large.

\subsection{Updating 3D Human Pose}

\noindent \textbf{Bone Local Encodings} 
Besides our point encodings taken from the canonical space, we also encode the angle and distance information from the sample points to each of the bone centers as additional information for our NeRF network \cite{su2021nerf}, as shown in \cref{fig:overview}. 

\vspace{0.2cm}
\noindent \textbf{Bounded Positional Embedding} 
In the conventional NeRF implementation, the positional embedding is applied to input vectors, where the periodic functions are used \cite{mildenhall2020nerf}, i.e. cosine and sine functions. 
In our case, we further bound the embedding by the distance between the sample points and the nearest vertex $u_1$ as bones should be less sensitive to distant pixels and more dependent on nearby ones. The bounded embedding is defined as:
\begin{equation}
\begin{split}
    \gamma(x) = &[x, D \times\sin(2^{0}x), D\times\cos(2^{0}x), \\
    &..., D\times\sin(2^{N}x), D\times\cos(2^{N}x)],
\end{split}
\end{equation}
\noindent where $x$ is the coordinate of a sample point, $D={\rm sigmoid}(\beta(x-u_{1}-\alpha))$ is the sigmoid cutoff controlled by predefined coefficients $\alpha$ and $\beta$.
Empirically, we set $\alpha=0.5$ and $\beta=100$ in our experiments.

\vspace{0.2cm}
\noindent\textbf{Loss Function} 
The loss function of our method is defined as the combination of a pixel-wise RGB loss and a pose regularization.
Given an input image $I$ and the reconstructed image $C$, the pixel-wise loss is the sum of $L_1$ differences at all sampled pixels $P$, which is formulated as:
\begin{equation}
    L_{\rm rgb} = \sum_{p \in P}|C(x_p) - I(x_p)|,
    \label{eq:RGB_loss}
\end{equation}
As the components in our pipeline are differentiable, the RGB loss can be back-propagated to optimize the 3D human poses. Moreover, the pose regularization loss aims to regularize the bone lengths across all the frames:
\begin{equation}
    L_{\rm reg} = \sum_b (B_b - \Tilde{B_b})^2,
\end{equation}
\noindent where $B_b$ is the $b^{th}$ bone vector and $\Tilde{B_b}$ is the average of estimated bone lengths. Our overall loss is the combination of the above losses: $L_{\rm all}=L_{\rm rgb} + \lambda L_{\rm reg}$, where $\lambda$ is set to $0.1$ in our experiments. 

\section{Experiments} 


\noindent \textbf{Datasets} 
\noindent- 3DPW \cite{3DPW} is an outdoor dataset with multi-person videos, which is used increasingly for 3D human pose and shape estimation. The videos are recorded outdoors where the object occlusions and inter-person occlusions happen frequently, which are suitable for evaluating the performance of different methods in multi-person scenarios. In particular, the video clip "crossStreets\_00" is used in evaluating different methods in this paper. No downsampling of the frame rate is applied. 

\noindent- Human 3.6M \cite{human36ionescu} is an indoor dataset with single-person videos, which is widely used in the human pose and shape estimation evaluation from video. The dataset contains four camera views and seven actors in total. The single-view (monocular) input and the test set (subjects S9 and S11) are used for quantitative evaluation following \cite{su2021nerf}.


\vspace{0.2cm}
\noindent \textbf{Evaluation Metrics} Following the existing rending-based 3D human modeling work \cite{su2021nerf}, Procrustes aligned MPJPE (P-MPJPE) is used to evaluate the accuracy of 3D human pose estimation. Joint regressors are used to convert from human mesh to 3D human pose following \cite{kocabas2020vibe, kolotouros2019spin}, where different regressors are used for Human 3.6M and 3DPW. 

\vspace{0.2cm}
\noindent \textbf{Implementation Details} 
\noindent - Network structure: The structure of temporal NeRF consists of two parts. For the network inferring the opacity field, 2 temporal convolutional layers and 2 MLP layers are used. The convolutional layer has stride length of 1 and window size of 3. For the network inferring the color field, 4 MLP layers are used. The output channel is set to 256 for both networks and the temporal length of input encodings is set to 5. In vertex encoding, we use a 32-dimension latent vector for each vertex.

\noindent - Training details: The whole pipeline is trained in an end-to-end manner, where the same training parameters are used for the temporal NeRF and 3D human pose updating. The training takes 400k iterations for 3D pose updates. The Adam optimizer is used with an initial learning rate of $5e-4$ and gradually decreased to $1e-5$ at the end. Each iteration of training takes 128 images and 32 2D locations are randomly sampled from each image within the average texture-based mask, and 96 sample points are selected along each ray. The training of our model takes 42 hours using 4x RTX 3090Ti GPUs, and the refined 3D human poses are obtained immediately after completing the training. Please refer to the supplementary material for more information. 

\begin{table}[tb]
	\footnotesize
	\centering
	\begin{tabular}{c|c|c}
		\cline{1-3}
		\rule{0pt}{2.6ex}
		\textbf{Method} & P-MPJPE & $\Delta$ initialization \\
		\cline{1-3}
		\rule{0pt}{2.6ex}
		Initialization (VIBE \cite{kocabas2020vibe})  & 59.9 & 0 \\
		HumanNeRF \cite{weng2022humannerf} & 65.5 & $\color{red}{\uparrow 5.6}$ \\
		A-NeRF \cite{su2021nerf} & \underline{60.2} & $\color{red}{\uparrow 0.3}$\\
		ORTexME (ours)  & \textbf{58.1} & $\color{blue}{\downarrow 1.8}$\\
		\cline{1-3}
	\end{tabular}
	\vspace{0.5em}
	\caption{3D human pose refinement performance evaluation on 3DPW testing set. $\Delta$ initialization indicates the change of the accuracy to the initial 3D pose. Best in \textbf{bold}, second best \underline{underlined}. }
	\label{tab:3dpw}
\end{table}

\subsection{Quantitative Comparisons}

The goal of our work is to improve the 3D human shape and pose estimation, not human neural rendering or novel-view synthesis. Thus, the evaluations are focused on the quality of 3D human shape and pose estimation. Note that 3D human shape and pose estimation (e.g., SMPL model \cite{loper2015smpl}) is different from the 3D human pose estimation that estimates the human skeleton only. As the existing consistency-based methods \cite{wandt2019repnet,zhang2020inference,cheng2022dual} solve the 3D human pose estimation problem, while the rendering-based methods solve the human shape and pose estimation like ours, therefore, we mainly compare to the existing rendering-based methods \cite{su2021nerf,weng2022humannerf}.

\noindent \textbf{Quantitative comparison on 3DPW}
To understand the 3D human pose performance of our method compared with the SOTA rendering-based 3D human pose modeling methods in multi-person settings, an evaluation on the 3DPW dataset is performed as shown in \cref{tab:3dpw}. 
We observe that the SOTA methods \cite{su2021nerf,weng2022humannerf} enlarge the 3D human pose errors by 5.6 - 0.3 P-MPJPE from the off-the-shelf method \cite{kocabas2020vibe}, instead of lowering the 3D human pose error. Our method is the first and the only one that achieves positive error reduction, where we achieve 1.8 P-MPJPE error reduction from the initialized poses in the challenging multi-person dataset. This evaluation demonstrates that our method is robust to occlusion and to incorrect segmentation that causes the failures of the SOTA methods.


\begin{table}[tb]
	\footnotesize
	\centering
	\begin{tabular}{c|c|c|c}
		\cline{1-4}
		\rule{0pt}{2.6ex}
		\textbf{Method} & S9 & S11 & Test set \\
		\cline{1-4}
		\rule{0pt}{2.6ex}
		Initialization (VIBE \cite{kocabas2020vibe}) & 42.24 & \underline{38.02} & 40.42 \\
		A-NeRF \cite{su2021nerf} & \underline{41.91} & 38.16 & \underline{40.29} \\
		ORTexME (ours) & \textbf{40.13} & \textbf{36.52} & \textbf{38.57} \\
		\cline{1-4}
	\end{tabular}
	\vspace{0.5em}
	\caption{Pose refinement evaluation on the Human3.6M testing set. P-MPJPE is used as the evaluation metric. Best in \textbf{bold}, second best \underline{underlined}. }
	\label{tab:h36m_test_set}
\end{table}

\noindent \textbf{Quantitative comparison on Human3.6M}
To further evaluate our method compared with the SOTA method (A-NeRF) \cite{su2021nerf}, another quantitative comparison is performed on the Human3.6M test set as shown in \cref{tab:h36m_test_set}. In particular, our method achieves 38.57 P-MPJPE where the initial 3D pose of the off-the-shelf method (VIBE \cite{kocabas2020vibe}) is of 40.42, resulting in 1.85 error reduction compared with A-NeRF's 0.13 reduction (40.29 P-MPJPE). 
A 14-fold improvement is achieved by our method compared to A-NeRF.
Note that the accuracy of A-NeRF in \cref{tab:h36m_test_set} may not match the reported performance in their paper. This is because we used predicted segmentation of Mask R-CNN \cite{he2017mask}, whereas the ground-truth segmentation is utilized in the paper of A-NeRF \cite{su2021nerf}. 

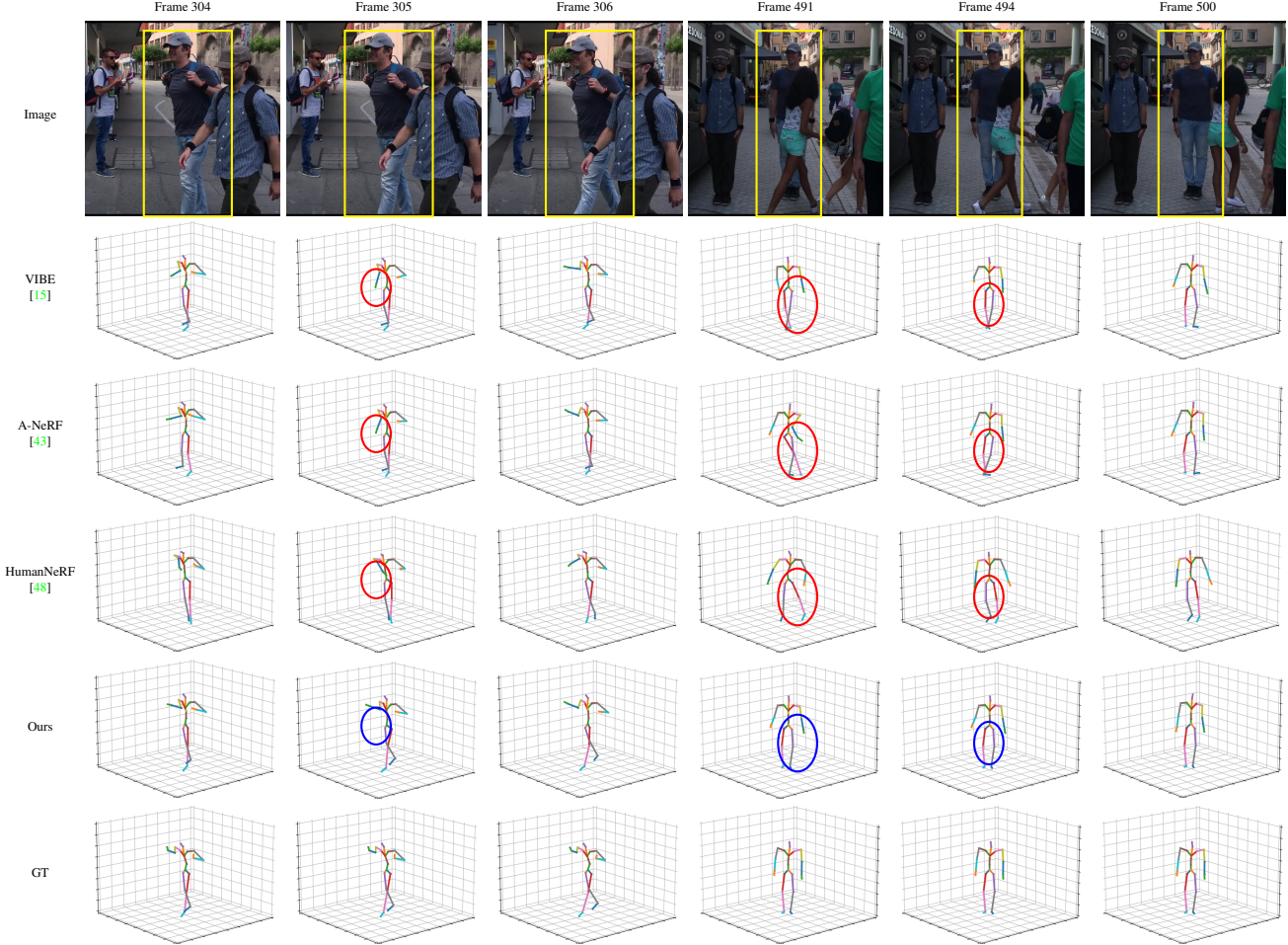
\begin{figure*}
    \centering
    
    \begin{minipage}[c]{0.06\linewidth}
    \centerline{\scriptsize{}}
	\vspace{0.3em}
	\end{minipage}
    \begin{minipage}[c]{0.150\linewidth}
    \centerline{\tiny{Frame 304}}
	\vspace{0.3em}
	\end{minipage}
	\begin{minipage}[c]{0.150\linewidth}
    \centerline{\tiny{Frame 305}}
	\vspace{0.3em}
	\end{minipage}
	\begin{minipage}[c]{0.150\linewidth}
    \centerline{\tiny{Frame 306}}
	\vspace{0.3em}
	\end{minipage}
	\begin{minipage}[c]{0.150\linewidth}
    \centerline{\tiny{Frame 491}}
	\vspace{0.3em}
	\end{minipage}
	\begin{minipage}[c]{0.150\linewidth}
    \centerline{\tiny{Frame 494}}
	\vspace{0.3em}
	\end{minipage}
	\begin{minipage}[c]{0.150\linewidth}
    \centerline{\tiny{Frame 500}}
	\vspace{0.3em}
	\end{minipage}
	
    \begin{minipage}[c]{0.06\linewidth}
    \centerline{\tiny{Image}}
	\vspace{0.3em}
	\end{minipage}
    \begin{minipage}[c]{0.150\linewidth}
	    \begin{tikzpicture}
	    \node[above right, inner sep=0] (image) at (0,0) {
		\includegraphics[trim={3.5\linewidth} {2.75\linewidth} {2\linewidth} {2.75\linewidth},clip=true,width=\linewidth]{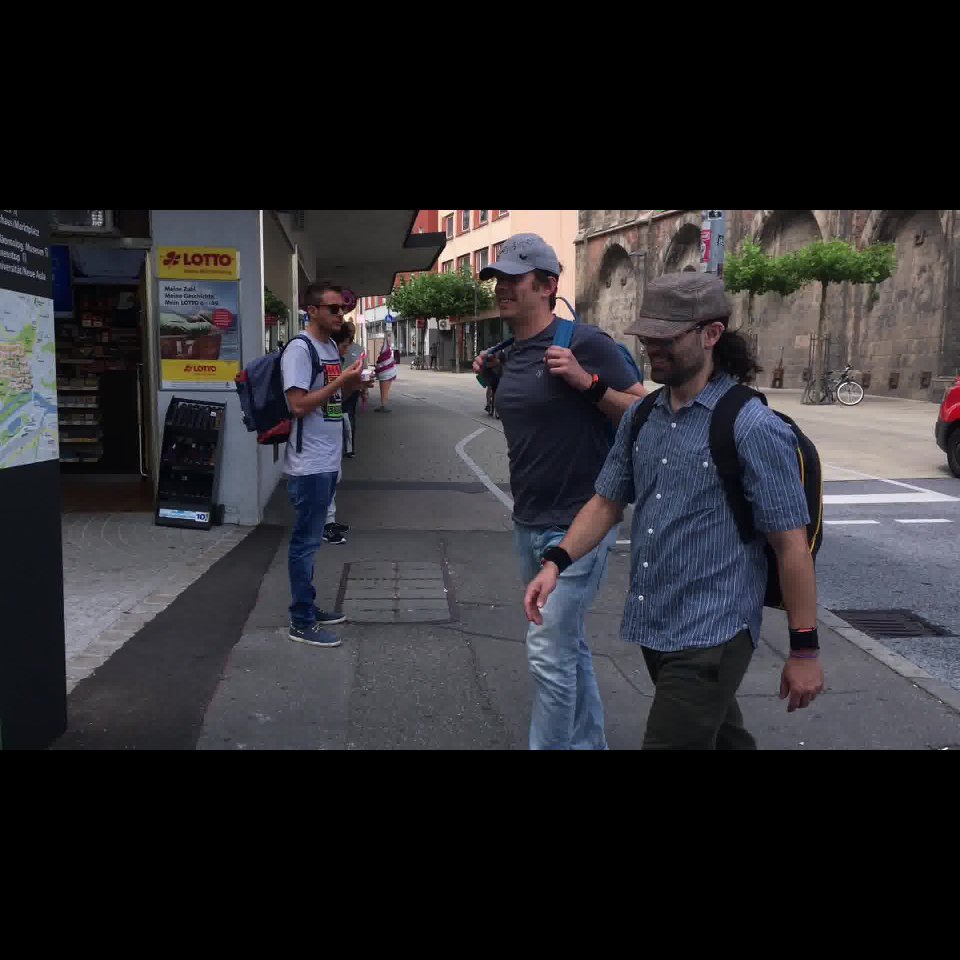}
		};
		\begin{scope}
		    [x={($0.1*(image.south east)$)},y={($0.1*(image.north west)$)}]
		    \draw[thick,yellow] (3,0) rectangle (7.5,9.5);
		\end{scope}
		\end{tikzpicture}
	\end{minipage}
	\begin{minipage}[c]{0.150\linewidth}
	    \begin{tikzpicture}
	    \node[above right, inner sep=0] (image) at (0,0) {
		\includegraphics[trim={3.5\linewidth} {2.75\linewidth} {2\linewidth} {2.75\linewidth},clip=true,width=\linewidth]{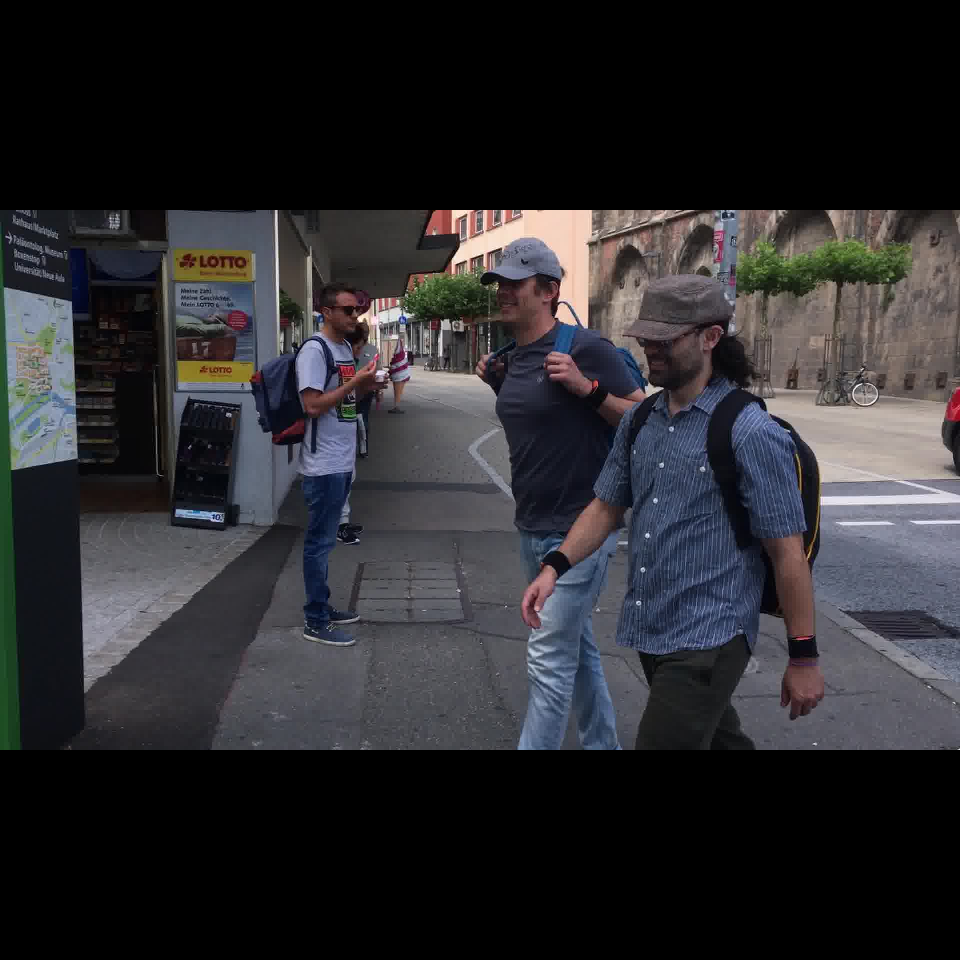}
		};
		\begin{scope}
		    [x={($0.1*(image.south east)$)},y={($0.1*(image.north west)$)}]
		    \draw[thick,yellow] (3,0) rectangle (7.5,9.5);
		\end{scope}
		\end{tikzpicture}
	\end{minipage}
	\begin{minipage}[c]{0.150\linewidth}
	    \begin{tikzpicture}
	    \node[above right, inner sep=0] (image) at (0,0) {
		\includegraphics[trim={3.5\linewidth} {2.75\linewidth} {2\linewidth} {2.75\linewidth},clip=true,width=\linewidth]{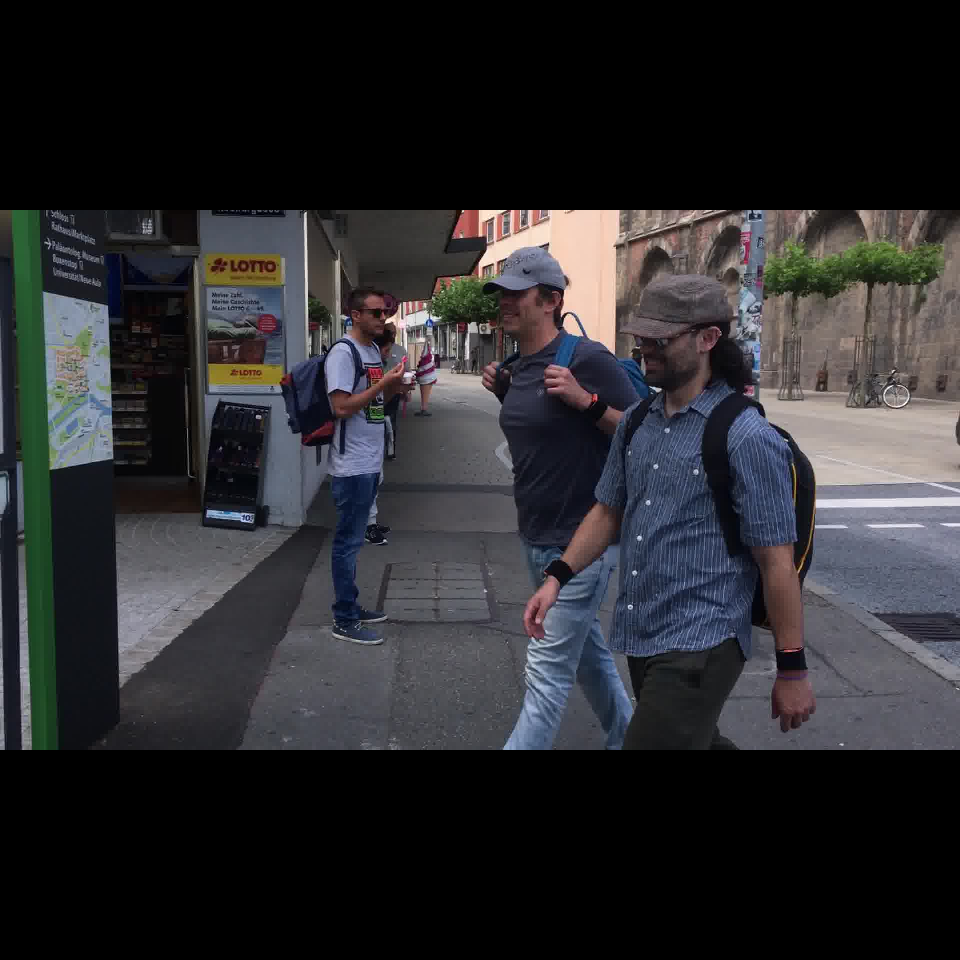}
		};
		\begin{scope}
		    [x={($0.1*(image.south east)$)},y={($0.1*(image.north west)$)}]
		    \draw[thick,yellow] (3,0) rectangle (7.5,9.5);
		\end{scope}
		\end{tikzpicture}
	\end{minipage}
	\begin{minipage}[c]{0.150\linewidth}
	    \begin{tikzpicture}
	    \node[above right, inner sep=0] (image) at (0,0) {
		\includegraphics[trim={4.5\linewidth} {2.75\linewidth} {1\linewidth} {2.75\linewidth},clip=true,width=\linewidth]{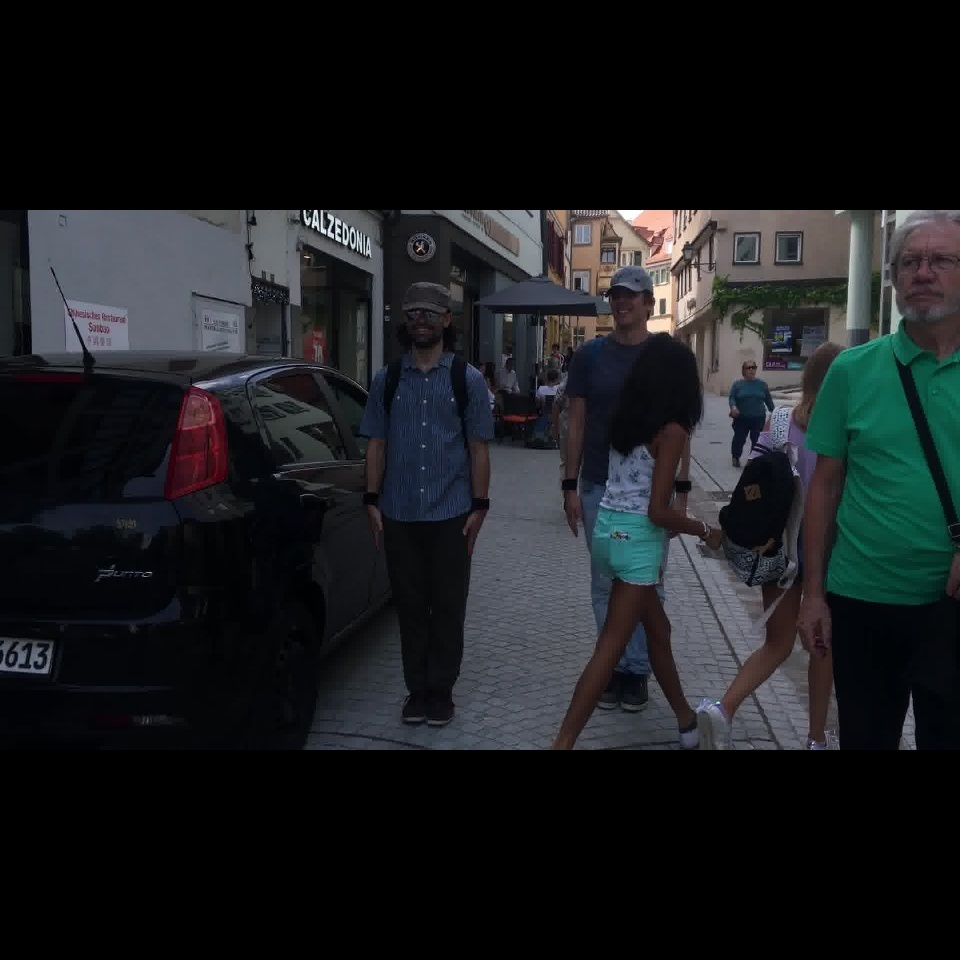}
		};
		\begin{scope}
		    [x={($0.1*(image.south east)$)},y={($0.1*(image.north west)$)}]
		    \draw[thick,yellow] (3.5,0) rectangle (6.8,9.5);
		\end{scope}
		\end{tikzpicture}
	\end{minipage}
	\begin{minipage}[c]{0.150\linewidth}
	    \begin{tikzpicture}
	    \node[above right, inner sep=0] (image) at (0,0) {
		\includegraphics[trim={4.5\linewidth} {2.75\linewidth} {1\linewidth} {2.75\linewidth},clip=true,width=\linewidth]{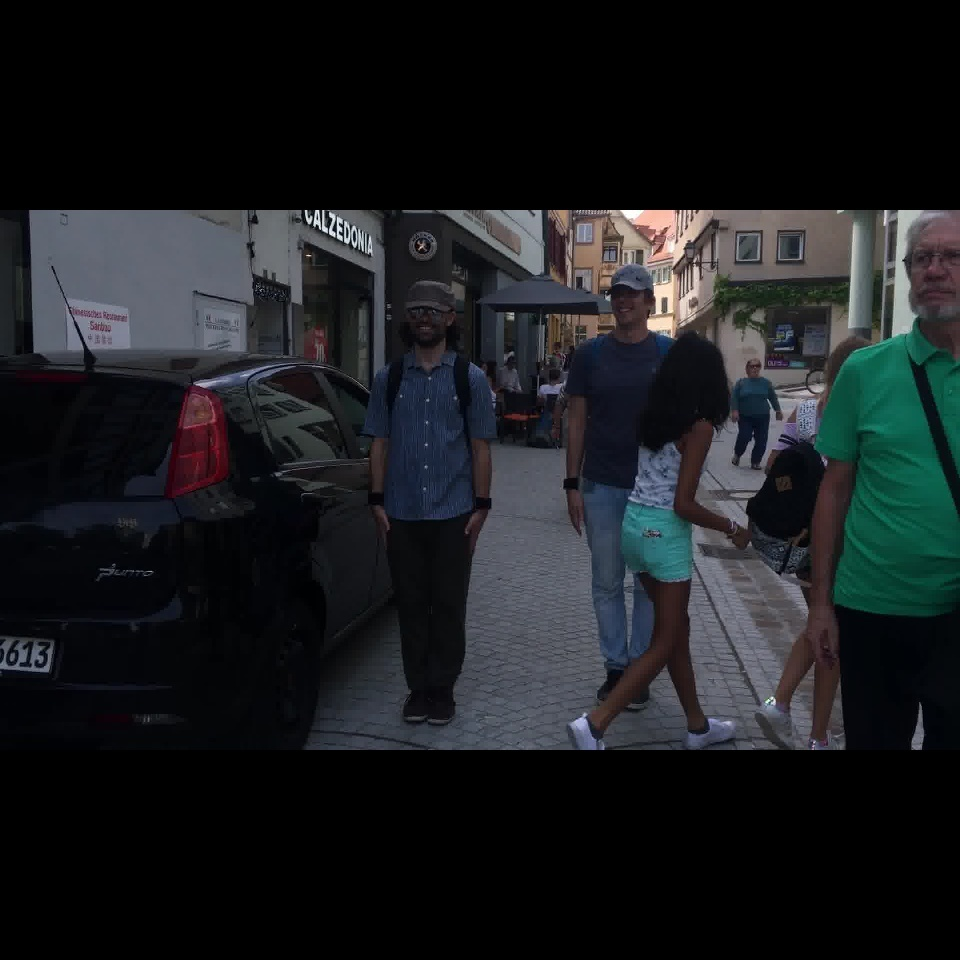}
		};
		\begin{scope}
		    [x={($0.1*(image.south east)$)},y={($0.1*(image.north west)$)}]
		    \draw[thick,yellow] (3.5,0) rectangle (6.8,9.5);
		\end{scope}
		\end{tikzpicture}
	\end{minipage}
	\begin{minipage}[c]{0.150\linewidth}
	    \begin{tikzpicture}
	    \node[above right, inner sep=0] (image) at (0,0) {
		\includegraphics[trim={4.5\linewidth} {2.75\linewidth} {1\linewidth} {2.75\linewidth},clip=true,width=\linewidth]{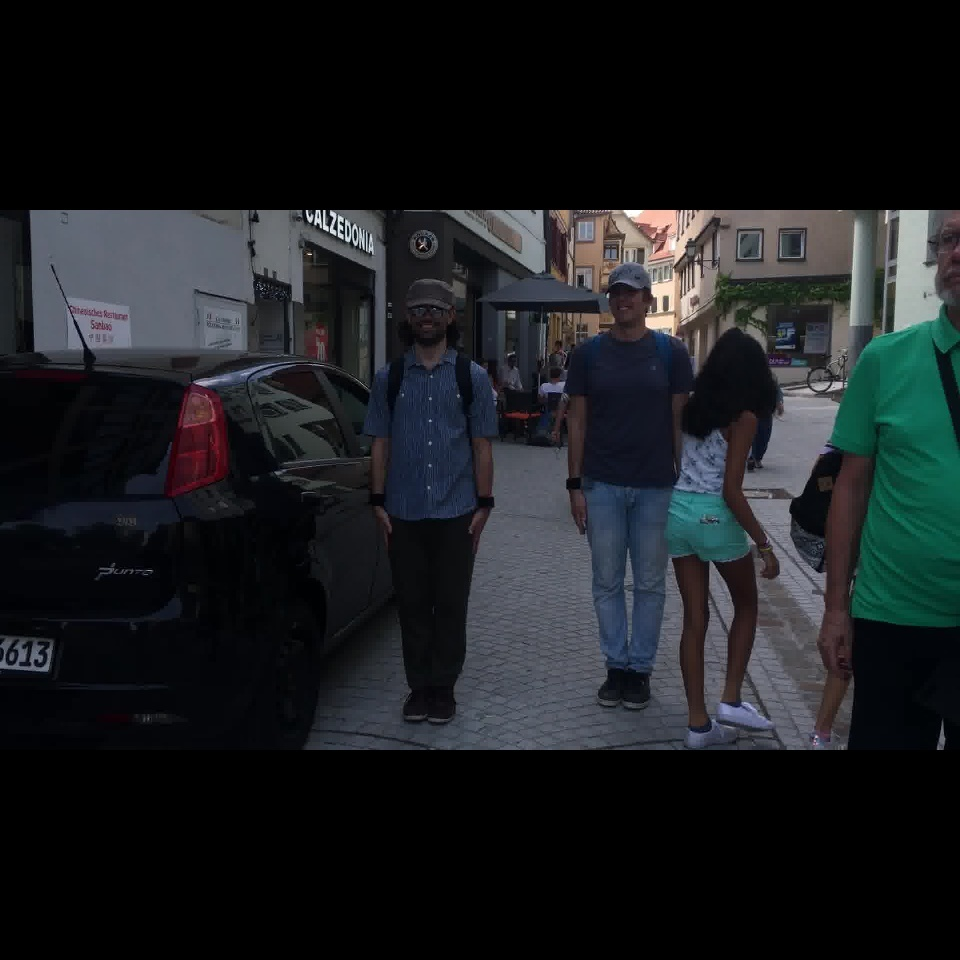}
		};
		\begin{scope}
		    [x={($0.1*(image.south east)$)},y={($0.1*(image.north west)$)}]
		    \draw[thick,yellow] (3.5,0) rectangle (6.8,9.5);
		\end{scope}
		\end{tikzpicture}
	\end{minipage}
	
	
	\begin{minipage}[c]{0.06\linewidth}
    \centerline{\tiny{VIBE}}
    \vspace{-0.6em}
    \centerline{\tiny{\cite{kocabas2020vibe}}}
	\vspace{0.3em}
	\end{minipage}
    \begin{minipage}[c]{0.150\linewidth}
		\includegraphics[width=\linewidth]{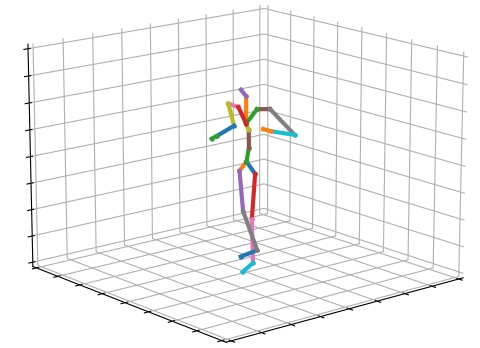}\vspace{0.1em}
	\end{minipage}
	\begin{minipage}[c]{0.150\linewidth}
	    \begin{tikzpicture}
	    \node[above right, inner sep=0] (image) at (0,0) {
		\includegraphics[width=\linewidth]{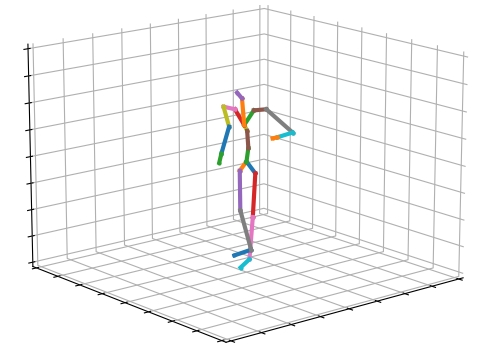}
		};
		\begin{scope}
		    [x={($0.1*(image.south east)$)},y={($0.1*(image.north west)$)}]
		    \draw[thick,red] (4.6, 5.3) ellipse (0.75 and 1.3);
		\end{scope}
		\end{tikzpicture}
	\end{minipage}
	\begin{minipage}[c]{0.150\linewidth}
		\includegraphics[width=\linewidth]{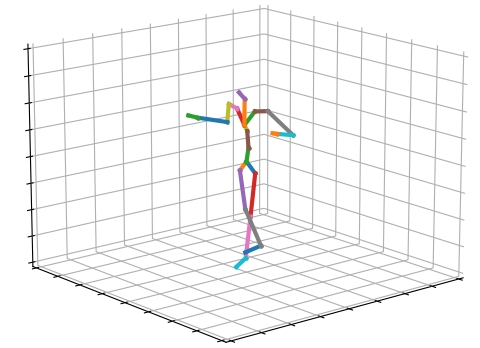}\vspace{0.1em}
	\end{minipage}
	\begin{minipage}[c]{0.150\linewidth}
	    \begin{tikzpicture}
	    \node[above right, inner sep=0] (image) at (0,0) {
		\includegraphics[width=\linewidth]{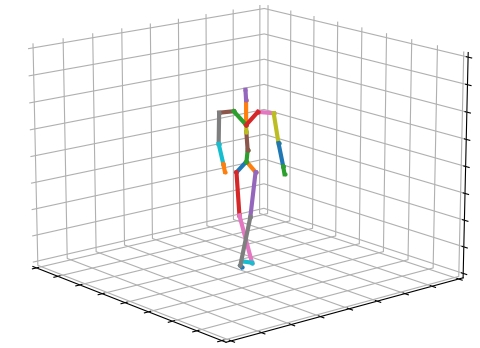}
		};
		\begin{scope}
		    [x={($0.1*(image.south east)$)},y={($0.1*(image.north west)$)}]
		    \draw[thick,red] (5.6, 4.1) ellipse (1 and 2);
		\end{scope}
		\end{tikzpicture}
	\end{minipage}
	\begin{minipage}[c]{0.150\linewidth}
	    \begin{tikzpicture}
	    \node[above right, inner sep=0] (image) at (0,0) {
		\includegraphics[width=\linewidth]{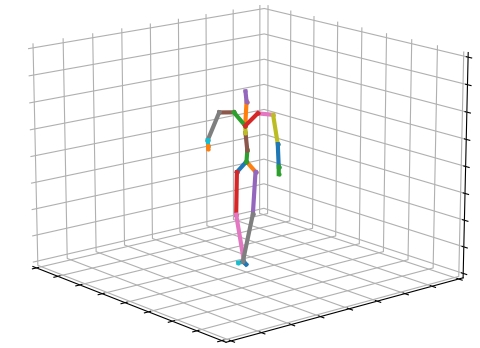}
		};
		\begin{scope}
		    [x={($0.1*(image.south east)$)},y={($0.1*(image.north west)$)}]
		    \draw[thick,red] (5.1, 4.1) ellipse (0.75 and 1.5);
		\end{scope}
		\end{tikzpicture}
	\end{minipage}
	\begin{minipage}[c]{0.150\linewidth}
		\includegraphics[width=\linewidth]{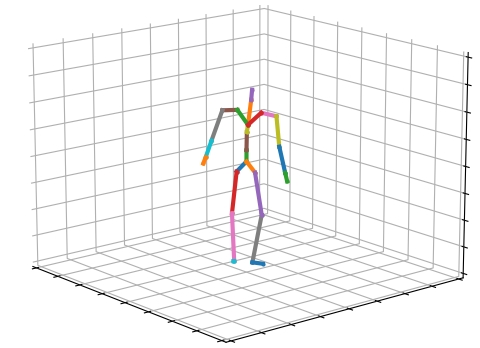}\vspace{0.1em}
	\end{minipage}
	
	\begin{minipage}[c]{0.06\linewidth}
    \centerline{\tiny{A-NeRF}}
    \vspace{-0.6em}
    \centerline{\tiny{\cite{su2021nerf}}}
	\vspace{0.3em}
	\end{minipage}
    \begin{minipage}[c]{0.150\linewidth}
		\includegraphics[width=\linewidth]{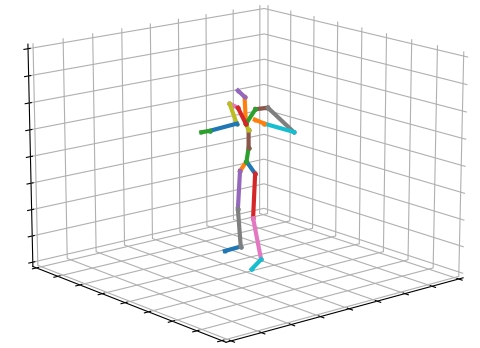}\vspace{0.1em}
	\end{minipage}
	\begin{minipage}[c]{0.150\linewidth}
	    \begin{tikzpicture}
	    \node[above right, inner sep=0] (image) at (0,0) {
		\includegraphics[width=\linewidth]{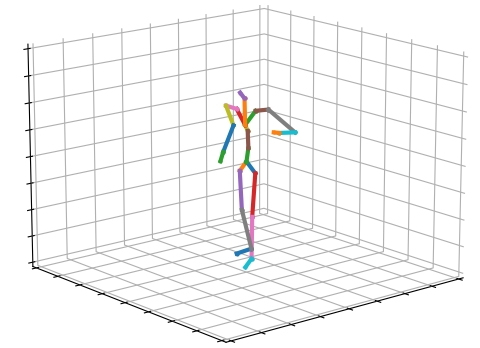}
		};
		\begin{scope}
		    [x={($0.1*(image.south east)$)},y={($0.1*(image.north west)$)}]
		    \draw[thick,red] (4.6, 5.3) ellipse (0.75 and 1.3);
		\end{scope}
		\end{tikzpicture}
	\end{minipage}
	\begin{minipage}[c]{0.150\linewidth}
		\includegraphics[width=\linewidth]{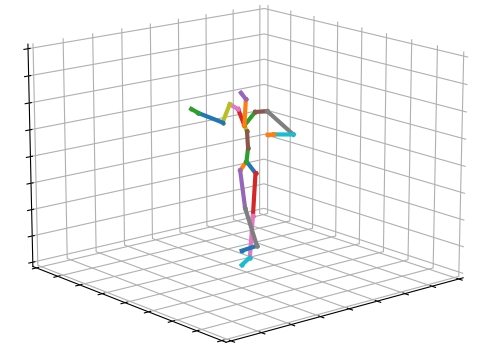}\vspace{0.1em}
	\end{minipage}
	\begin{minipage}[c]{0.150\linewidth}
		\begin{tikzpicture}
	    \node[above right, inner sep=0] (image) at (0,0) {
		\includegraphics[width=\linewidth]{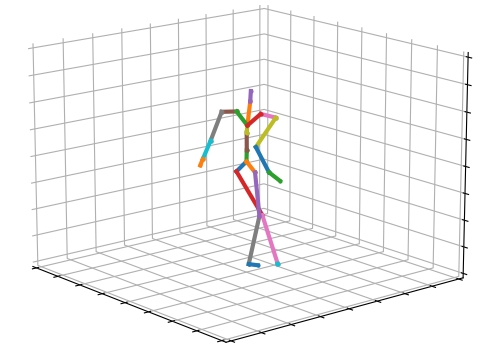}
		};
		\begin{scope}
            [x={($0.1*(image.south east)$)},y={($0.1*(image.north west)$)}]
            \draw[thick,red] (5.6, 4.1) ellipse (1 and 2);
        \end{scope}
		\end{tikzpicture}
	\end{minipage}
	\begin{minipage}[c]{0.150\linewidth}
	    \begin{tikzpicture}
	    \node[above right, inner sep=0] (image) at (0,0) {
		\includegraphics[width=\linewidth]{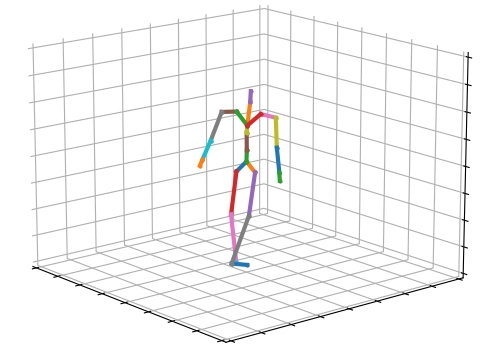}
		};
		\begin{scope}
		    [x={($0.1*(image.south east)$)},y={($0.1*(image.north west)$)}]
		    \draw[thick,red] (5.1, 4.1) ellipse (0.75 and 1.5);
		\end{scope}
		\end{tikzpicture}
	\end{minipage}
	\begin{minipage}[c]{0.150\linewidth}
		\includegraphics[width=\linewidth]{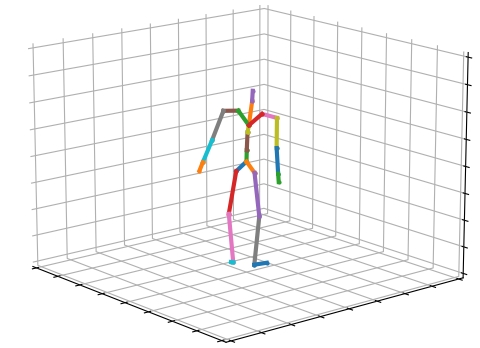}\vspace{0.1em}
	\end{minipage}

	\begin{minipage}[c]{0.06\linewidth}
    \centerline{\tiny{HumanNeRF}}
    \vspace{-0.6em}
    \centerline{\tiny{\cite{weng2022humannerf}}}
	\vspace{0.3em}
	\end{minipage}
    \begin{minipage}[c]{0.150\linewidth}
		\includegraphics[width=\linewidth]{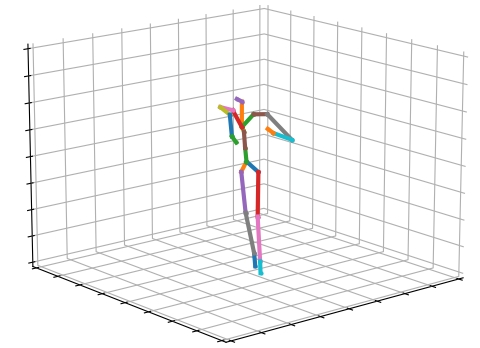}\vspace{0.1em}
	\end{minipage}
	\begin{minipage}[c]{0.150\linewidth}
	    \begin{tikzpicture}
	    \node[above right, inner sep=0] (image) at (0,0) {
		\includegraphics[width=\linewidth]{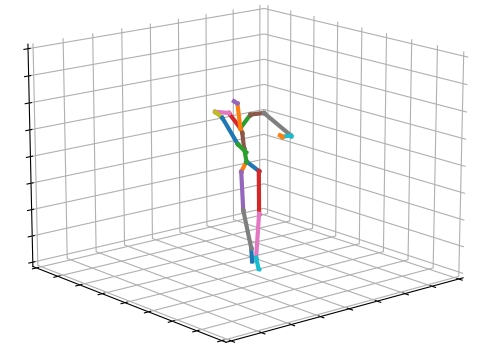}
		};
		\begin{scope}
		    [x={($0.1*(image.south east)$)},y={($0.1*(image.north west)$)}]
		    \draw[thick,red] (4.6, 5.3) ellipse (0.75 and 1.3);
		\end{scope}
		\end{tikzpicture}
	\end{minipage}
	\begin{minipage}[c]{0.150\linewidth}
		\includegraphics[width=\linewidth]{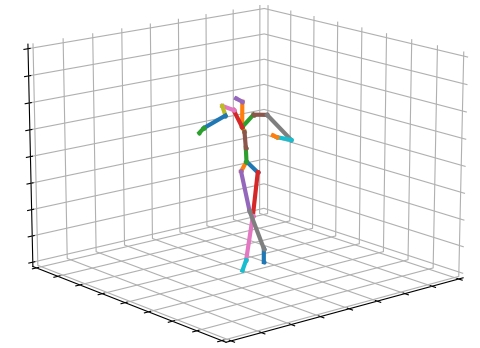}\vspace{0.1em}
	\end{minipage}
	\begin{minipage}[c]{0.150\linewidth}
		\begin{tikzpicture}
	    \node[above right, inner sep=0] (image) at (0,0) {
		\includegraphics[width=\linewidth]{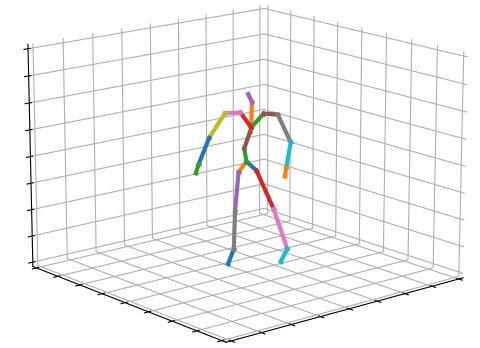}
		};
		\begin{scope}
            [x={($0.1*(image.south east)$)},y={($0.1*(image.north west)$)}]
            \draw[thick,red] (5.6, 4.1) ellipse (1 and 2);
        \end{scope}
		\end{tikzpicture}
	\end{minipage}
	\begin{minipage}[c]{0.150\linewidth}
	    \begin{tikzpicture}
	    \node[above right, inner sep=0] (image) at (0,0) {
		\includegraphics[width=\linewidth]{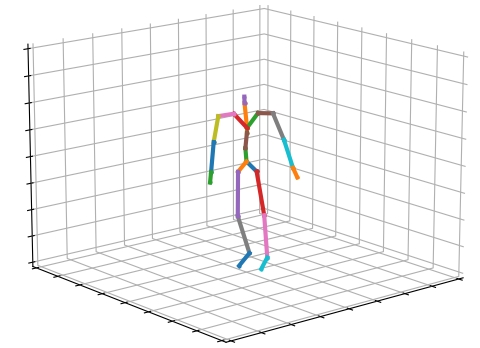}
		};
		\begin{scope}
		    [x={($0.1*(image.south east)$)},y={($0.1*(image.north west)$)}]
		    \draw[thick,red] (5.1, 4.1) ellipse (0.75 and 1.5);
		\end{scope}
		\end{tikzpicture}
	\end{minipage}
	\begin{minipage}[c]{0.150\linewidth}
		\includegraphics[width=\linewidth]{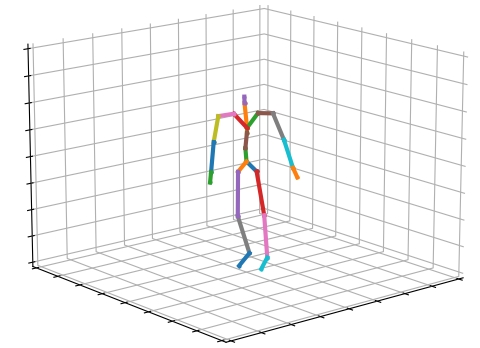}\vspace{0.1em}
	\end{minipage}
	
	\begin{minipage}[c]{0.06\linewidth}
    \centerline{\tiny{Ours}}
	\vspace{0.3em}
	\end{minipage}
    \begin{minipage}[c]{0.150\linewidth}
		\includegraphics[width=\linewidth]{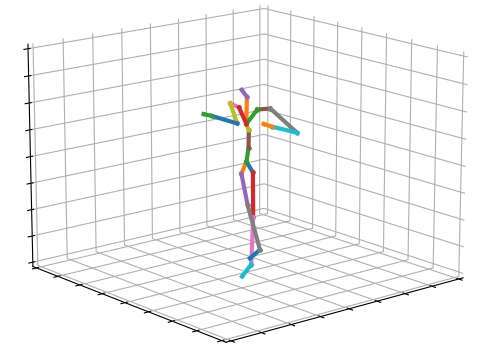}\vspace{0.1em}
	\end{minipage}
	\begin{minipage}[c]{0.150\linewidth}
	    \begin{tikzpicture}
	    \node[above right, inner sep=0] (image) at (0,0) {
		\includegraphics[width=\linewidth]{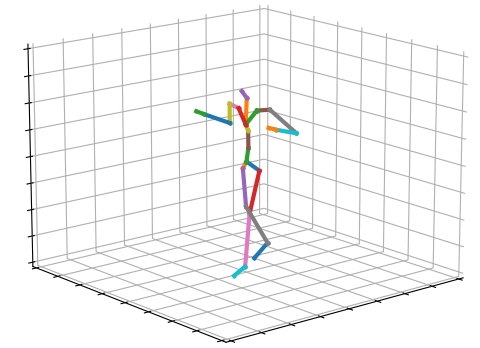}
		};
		\begin{scope}
		    [x={($0.1*(image.south east)$)},y={($0.1*(image.north west)$)}]
		    \draw[thick,blue] (4.6, 5.3) ellipse (0.75 and 1.3);
		\end{scope}
		\end{tikzpicture}
	\end{minipage}
	\begin{minipage}[c]{0.150\linewidth}
		\includegraphics[width=\linewidth]{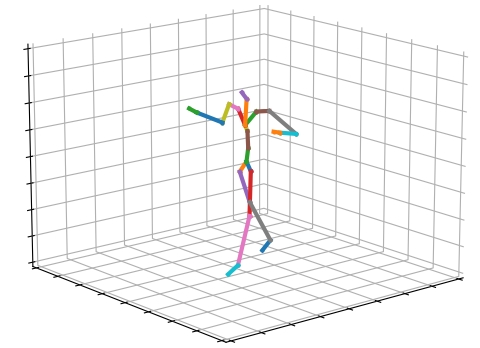}\vspace{0.1em}
	\end{minipage}
	\begin{minipage}[c]{0.150\linewidth}
	    \begin{tikzpicture}
	    \node[above right, inner sep=0] (image) at (0,0) {
		\includegraphics[width=\linewidth]{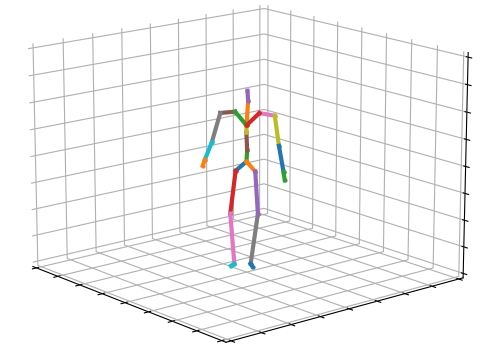}
		};
		\begin{scope}
		    [x={($0.1*(image.south east)$)},y={($0.1*(image.north west)$)}]
		    \draw[thick,blue] (5.6, 4.1) ellipse (1 and 2);
		\end{scope}
		\end{tikzpicture}
	\end{minipage}
	\begin{minipage}[c]{0.150\linewidth}
	    \begin{tikzpicture}
	    \node[above right, inner sep=0] (image) at (0,0) {
		\includegraphics[width=\linewidth]{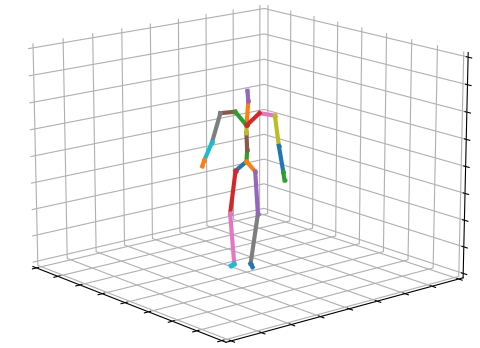}
		};
		\begin{scope}
		    [x={($0.1*(image.south east)$)},y={($0.1*(image.north west)$)}]
		    \draw[thick,blue] (5.1, 4.1) ellipse (0.75 and 1.5);
		\end{scope}
		\end{tikzpicture}
	\end{minipage}
	\begin{minipage}[c]{0.150\linewidth}
		\includegraphics[width=\linewidth]{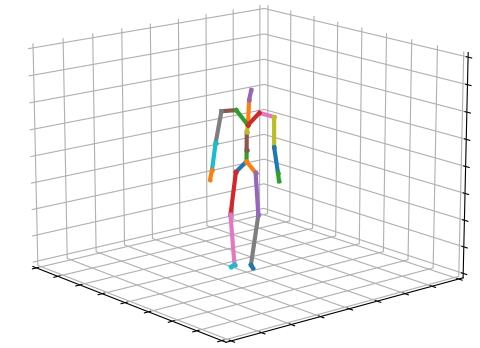}\vspace{0.1em}
	\end{minipage}
	
	\begin{minipage}[c]{0.06\linewidth}
    \centerline{\tiny{GT}}
	\vspace{0.3em}
	\end{minipage}
    \begin{minipage}[c]{0.150\linewidth}
		\includegraphics[width=\linewidth]{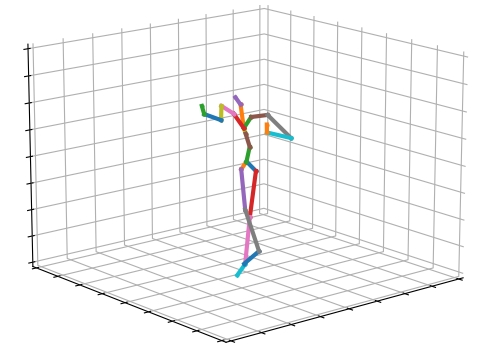}\vspace{0.1em}
	\end{minipage}
	\begin{minipage}[c]{0.150\linewidth}
		\includegraphics[width=\linewidth]{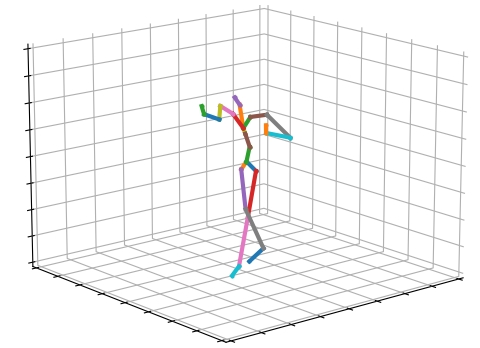}\vspace{0.1em}
	\end{minipage}
	\begin{minipage}[c]{0.150\linewidth}
		\includegraphics[width=\linewidth]{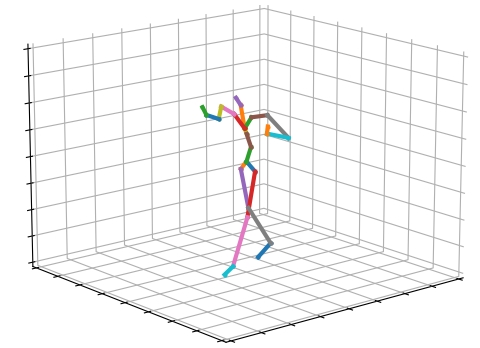}\vspace{0.1em}
	\end{minipage}
	\begin{minipage}[c]{0.150\linewidth}
		\includegraphics[width=\linewidth]{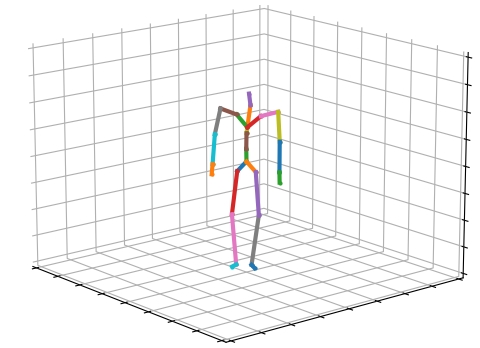}\vspace{0.1em}
	\end{minipage}
	\begin{minipage}[c]{0.150\linewidth}
		\includegraphics[width=\linewidth]{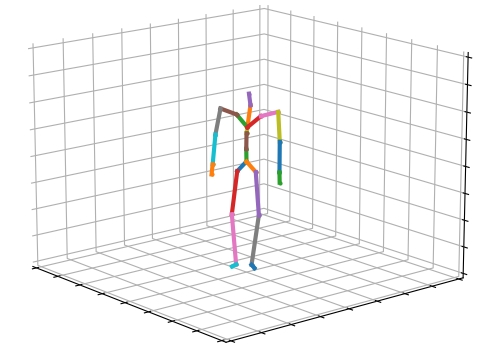}\vspace{0.1em}
	\end{minipage}
	\begin{minipage}[c]{0.150\linewidth}
		\includegraphics[width=\linewidth]{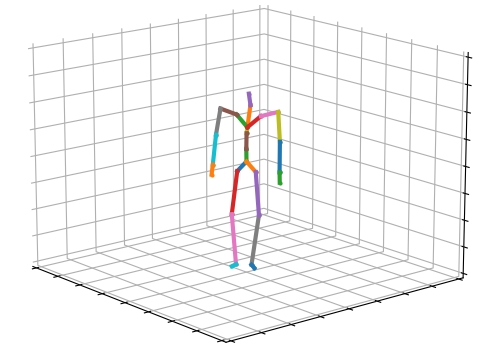}\vspace{0.1em}
	\end{minipage}

    \caption{Qualitative comparison with the SOTA methods \cite{su2021nerf,weng2022humannerf} on 3DPW dataset. The target person is labeled with yellow bounding box in the first row, and all the predicted and ground-truth human poses are for the target person. The major errors of the initial poses (VIBE \cite{kocabas2020vibe}) and the refined ones by A-NeRF are highlighted with red circles, where our improvements are highlighted with blue circles.}
    
    
    \label{fig:qualitative_compare_3dpw}
\end{figure*}

\subsection{Qualitative Comparisons}

The output of our method is 3D human shape and pose (e.g., SMPL model), and the result of 3D human poses are provided to be consistent with the visualization of the existing method \cite{su2021nerf}. \cref{fig:qualitative_compare_3dpw} shows the qualitative comparisons to the SOTA rendering-based methods \cite{su2021nerf,weng2022humannerf} on the 3DPW dataset, where it is observed that our method can effectively refine the 3D human poses and be close to the ground truth, while the SOTA methods fail to produce reasonable 3D human poses. 
Frame 491 shows an inter-person occlusion example, where the initial pose is inaccurate and the refinements of the existing methods \cite{su2021nerf,weng2022humannerf} make it much worse because they are sensitive to wrong segmentation and occlusions. On the contrary, our method generates accurate poses benefiting from the temporal NeRF and masking modules. 
The comparisons in \cref{fig:qualitative_compare_3dpw} show that our method can effectively perform 3D human pose refinement in challenging multi-person scenarios. 

\subsection{Ablation Studies}

\begin{table}[t]
	\footnotesize
	\centering
	\begin{tabular}{c|c|c}
		\cline{1-3}
		\rule{0pt}{2.6ex}
		\textbf{Method} & P-MPJPE & $\Delta$ \\
		\cline{1-3}
		\rule{0pt}{2.6ex}
		Initialization (VIBE) & 59.9 & 0\\
		Baseline & 60.4 & $\color{red}{\uparrow 0.5}$\\
		+ Masking  & 59.4 & $\color{blue}{\downarrow 1.0}$\\
		+ Point encoding  & 59.0 & $\color{blue}{\downarrow 0.4}$\\
		+ Vertex encoding  & 58.8 & $\color{blue}{\downarrow 0.2}$\\
		+ Temporal NeRF  & 58.1 & $\color{blue}{\downarrow 0.7}$\\
		\cline{1-3}
	\end{tabular}
	\vspace{0.5em}
	\caption{Ablation studies on the 3DPW dataset. $\Delta$ indicates the change of the accuracy between the current row and the row above. Best in \textbf{bold}, second best \underline{underlined}.}
	\label{tab:ablation_3dpw}
\end{table}

We evaluate the major components in our method on the multi-person 3DPW dataset as displayed in \cref{tab:ablation_3dpw}. 
The baseline means our pipeline without any of the newly proposed components, where the bone local encoding is adopted from A-NeRF \cite{su2021nerf} (i.e., angle and length from sample points to each of the bone centers). On top of the baseline, we add each of the major components and evaluate the 3D human pose accuracy. 
We observe that all the components effectively improve the 3D human pose accuracy. Specifically, the masking and our temporal NeRF modules make the most contribution to the total error reduction. In particular, the two modules combined account for 73.9\% of the total error reduction (i.e., $1.7/2.3$) on the challenging multi-person 3DPW dataset, which is higher than the 65.1\% of the two modules contributed to the total error reduction on the single-person Human3.6M dataset (\cref{tab:ablation_h36m}). The higher percentage of contribution 
demonstrates that handling occlusion and wrong human segmentation are crucial to improving 3D human pose accuracy in the complex multi-person scenarios, and validates the effectiveness of our masking and our temporal NeRF components. 

\begin{table}[t]
	\footnotesize
	\centering
	\begin{tabular}{c|c|c}
		\cline{1-3}
		\rule{0pt}{2.6ex}
		\textbf{Method} & P-MPJPE & $\Delta$\\
		\cline{1-3}
		\rule{0pt}{2.6ex}
		Initialization (VIBE)  & 42.24 & 0\\
		Baseline & 43.71 & $\color{red}{\uparrow 1.47}$\\
		+ Masking  & 42.17 & $\color{blue}{\downarrow 1.54}$\\
		+ Point encoding  & 41.49 & $\color{blue}{\downarrow 0.68}$\\
		+ Vertex encoding  & 40.92 & $\color{blue}{\downarrow 0.57}$\\
		+ Temporal NeRF  & 40.13 & $\color{blue}{\downarrow 0.79}$\\
		\cline{1-3}
	\end{tabular}
	\vspace{0.5em}
	\caption{Ablation studies on the Human3.6M test set (subject S9). $\Delta$ indicates the change in the accuracy between the current row and row above. Best in \textbf{bold}, second best \underline{underlined}.}
	\label{tab:ablation_h36m}
\end{table}

\noindent \textbf{Ablation study on Human3.6M}
Additional ablation studies are performed on the Human3.6M dataset, as shown in \cref{tab:ablation_h36m}. The baseline and the setting are the same as in \cref{tab:ablation_3dpw}. We observe that each component in our method helps to reduce the 3D human pose error. Among the four major components, the masking and the temporal NeRF modules reduce the error the most, with 1.54 and 0.79 P-MPJPE reduction respectively, which demonstrate the effectiveness of both components in terms of handling occlusion and human segmentation inaccuracies.

\begin{table}[t]
	\footnotesize
	\centering
	\begin{tabular}{c|c|c|c|c}
		\cline{1-5}
		\rule{0pt}{2.6ex}
		\textbf{Method} & SPIN \cite{kolotouros2019spin} & VIBE \cite{kocabas2020vibe} & TCMR \cite{choi2021beyond} & Avg.\\
		\cline{1-5}
		\rule{0pt}{2.6ex}
		Initialization & 43.65 & 42.24 & \underline{41.97} & 42.62\\
		A-NeRF \cite{su2021nerf} & \underline{41.35} & \underline{41.91} & 42.76 & \underline{42.01}\\
		ORTexME (ours)  & \textbf{40.81} & \textbf{40.13} & \textbf{40.64} & \textbf{40.53}\\
		\cline{1-5}
	\end{tabular}
	\vspace{0.5em}
	\caption{3D human pose refinement performance evaluation on the Human3.6M testing set (subject S9) with different off-the-shelf methods for initializations. Best in \textbf{bold}, second best \underline{underlined}. }
	\label{tab:different_initial_methods_h36m}
\end{table}

\noindent \textbf{Evaluation with different initializations}
\cref{tab:different_initial_methods_h36m} is an evaluation with three different off-the-shelf 3D human pose and shape estimation methods \cite{kolotouros2019spin,kocabas2020vibe,choi2021beyond} on the Human3.6M dataset to validate whether our method can consistently improve the initial 3D human poses from different methods.
In \cref{tab:different_initial_methods_h36m}, the first row shows the 3D human pose accuracy of the three off-the-shelf methods, the second and third rows show the accuracy of the SOTA method (A-NeRF) \cite{su2021nerf} and ours. It is observed that A-NeRF fails to consistently reduce the 3D human pose error, where A-NeRF enlarges the error to 42.76 from 41.97 of TCMR \cite{choi2021beyond}.
In contrast, our method consistently reduces the error across all three different off-the-shelf methods, including TCMR \cite{choi2021beyond} where temporal information is already utilized. On average, our error reduction is 2.09 P-MPJPE compared with that of A-NeRF, which is only 0.61 P-MPJPE, which illustrates our method is applicable to various off-the-shelf 3D pose and shape estimation methods where our method consistently improves the accuracy of 3D pose estimation and reduces the error.



\section{Limitation and Failure Cases}

\begin{figure}
    \centering

    \begin{minipage}[c]{0.32\linewidth}
        \centerline{\tiny{Image}}
	    \vspace{0.3em}
	\end{minipage}
	\begin{minipage}[c]{0.32\linewidth}
        \centerline{\tiny{Ours}}
	    \vspace{0.3em}
	\end{minipage}
	\begin{minipage}[c]{0.32\linewidth}
        \centerline{\tiny{GT}}
	    \vspace{0.3em}
	\end{minipage}

    \begin{minipage}[c]{0.32\linewidth}
	    \begin{tikzpicture}
	    \node[above right, inner sep=0] (image) at (0,0) {
		\includegraphics[trim={0cm} {7cm} {0cm} {2.5cm},clip=true,width=\linewidth]{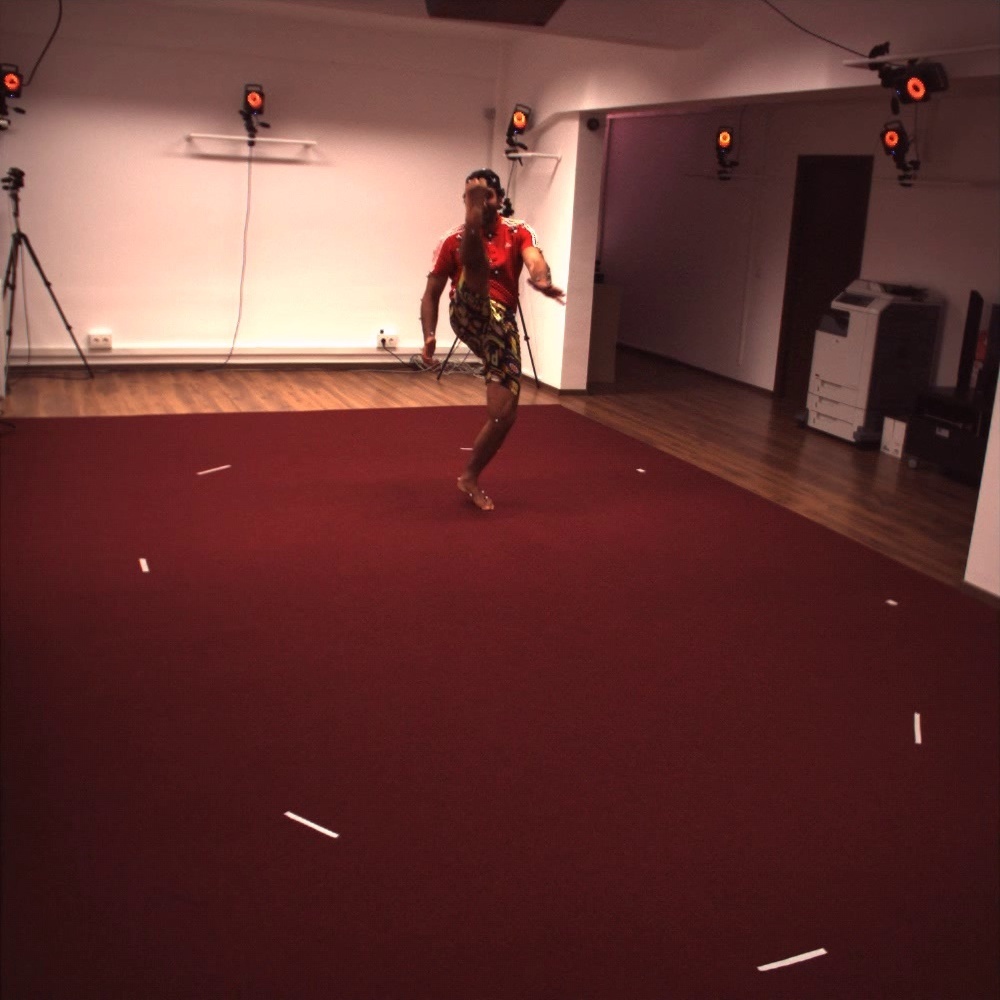}
		};
		\end{tikzpicture}
	\end{minipage}
    \begin{minipage}[c]{0.32\linewidth}
	    \begin{tikzpicture}
	    \node[above right, inner sep=0] (image) at (0,0) {
		\includegraphics[trim={0cm} {2.5cm} {0cm} {2.5cm},clip=true,width=\linewidth]{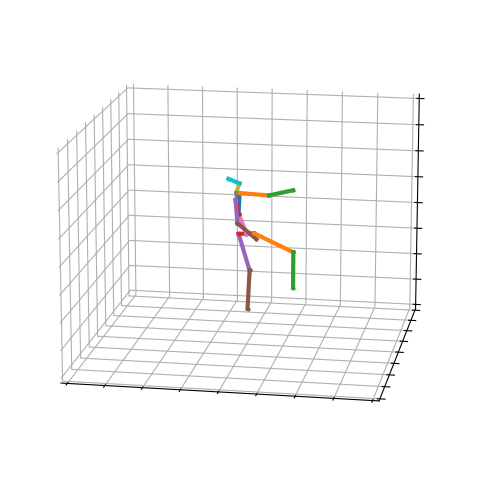}
		};
		\end{tikzpicture}
	\end{minipage}
    \begin{minipage}[c]{0.32\linewidth}
	    \begin{tikzpicture}
	    \node[above right, inner sep=0] (image) at (0,0) {
		\includegraphics[trim={0cm} {2.5cm} {0cm} {2.5cm},clip=true,width=\linewidth]{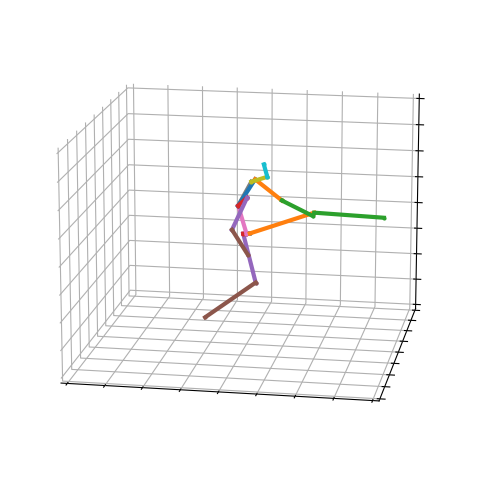}
		};
		\end{tikzpicture}
	\end{minipage}

    \begin{minipage}[c]{0.32\linewidth}
	    \begin{tikzpicture}
	    \node[above right, inner sep=0] (image) at (0,0) {
		\includegraphics[trim={0cm} {7cm} {0cm} {2.5cm},clip=true,width=\linewidth]{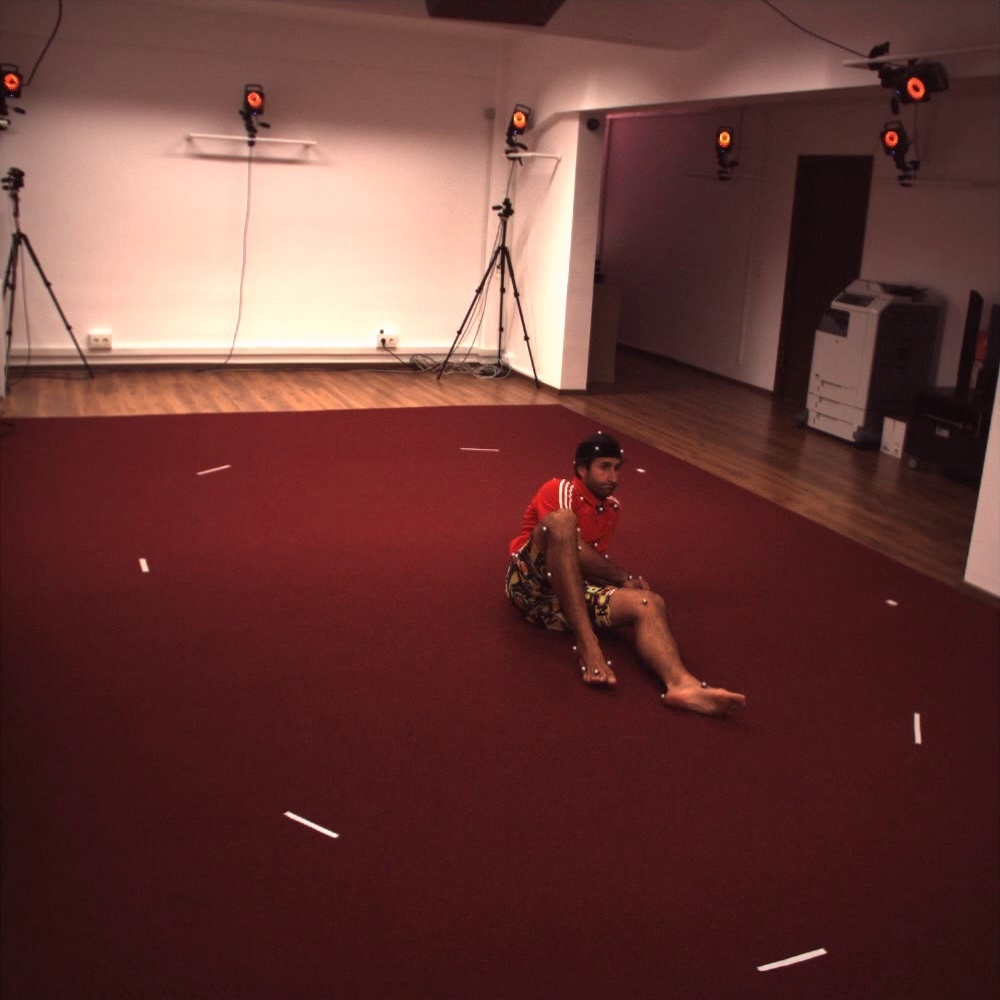}
		};
		\end{tikzpicture}
	\end{minipage}
    \begin{minipage}[c]{0.32\linewidth}
	    \begin{tikzpicture}
	    \node[above right, inner sep=0] (image) at (0,0) {
		\includegraphics[trim={0cm} {2.2cm} {0cm} {2.5cm},clip=true,width=\linewidth]{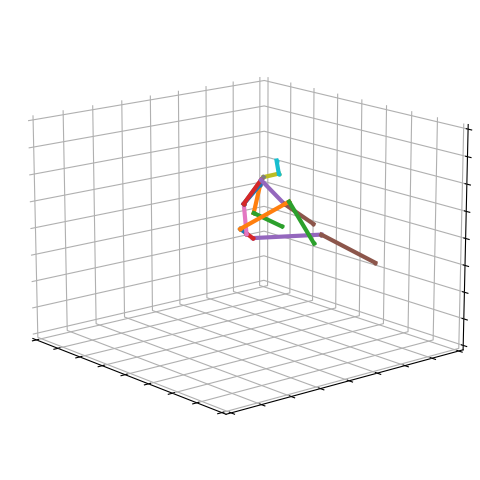}
		};
		\end{tikzpicture}
	\end{minipage}
    \begin{minipage}[c]{0.32\linewidth}
	    \begin{tikzpicture}
	    \node[above right, inner sep=0] (image) at (0,0) {
		\includegraphics[trim={0cm} {2.2cm} {0cm} {2.5cm},clip=true,width=\linewidth]{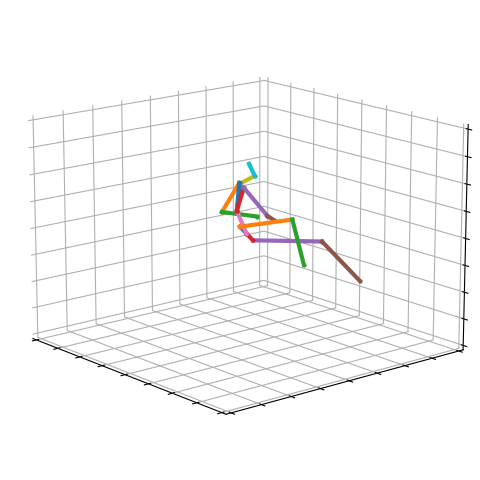}
		};
		\end{tikzpicture}
	\end{minipage}

	\vspace{0.2cm}
    \caption{Example results of the failure cases. Upper: The fast motion causes degraded image quality and insufficient temporal information for pose refinement. Lower: Refined pose is temporally smooth but still different from GT due to persistent occlusion.}
    \vspace{-0.2em}

    \label{fig:failure}
\end{figure}

We include two types of failure cases of our method in \cref{fig:failure}. The first type of failure is caused by image degradation (top example of \cref{fig:failure}). As our method relies on RGB loss to refine 3D human pose, if the observed person in the image has image degradation due to motion blur or noise. This may cause the loss computation to be ineffective, if the image degradation happens together with fast motion (e.g., action happens within 4-5 frames), the temporal information may not be able to capture the abrupt motion. 
Persistent occlusion is the reason for the second type of failure as shown at the bottom of \cref{fig:failure}. When occlusion is persistent, the model is able to produce temporally smooth prediction, which may differ from the ground truth. Moreover, the current training time is relatively slow, speeding up the training is a future direction to be explored \cite{reiser2021kilonerf,chen2022geometry}. 

\section{Conclusion}

We presented ORTexME to refine 3D human poses from off-the-shelf methods. Unlike existing rendering-based methods that are sensitive to occlusion and wrong human segmentation, our ORTexME is able to effectively improve the 3D human pose accuracy in single-person settings and in more challenging multi-person settings. Benefiting from our novel masking and temporal NeRF modules, we are the first to achieve effective 3D pose error reduction on the complex multi-person 3DPW dataset, where the SOTA methods fail to reduce but enlarge the 3D pose errors. 




{\small
\bibliographystyle{ieee_fullname}
\bibliography{egbib}
}

\end{document}